\documentclass[11pt,table]{article}
\pdfoutput=1

\usepackage[utf8]{inputenc} % allow utf-8 input
\usepackage[T1]{fontenc}    % use 8-bit T1 fonts
\usepackage[in]{fullpage}
\usepackage{booktabs}       % professional-quality tables
\usepackage{amsfonts}       % blackboard math symbols
\usepackage{nicefrac}       % compact symbols for 1/2, etc.
\usepackage{microtype}      % microtypography
\usepackage{amsmath,amssymb,amsthm}
\usepackage{mathtools}
\usepackage{mathrsfs}
\usepackage{thmtools}
\usepackage{thm-restate}
\usepackage{comment}
\usepackage{parskip}
\usepackage{siunitx}
\usepackage{etoolbox}
\usepackage{enumitem}
\usepackage{graphicx}
\usepackage{subcaption}
\usepackage{tabularx}
\usepackage{xcolor} 
\usepackage[numbers,sort]{natbib}
\usepackage{hyperref}
\usepackage{longtable}
\usepackage{afterpage}
\usepackage{footmisc}
\usepackage{booktabs}
\usepackage{graphicx}
\usepackage{amsfonts}
\usepackage{amsmath}
\usepackage{amsthm}
\usepackage{mathtools}
\usepackage{xcolor}
\usepackage{enumerate}
\usepackage{wrapfig}
\usepackage{subcaption}
\usepackage{gensymb}

\usepackage{thmtools}
\usepackage{thm-restate}
\usepackage{cleveref}

\newtheorem{exmp}{Example}[section]

\newtheorem*{lem*}{Lemma}

\DeclareMathOperator*{\argmin}{arg\,min}
\newcommand{\RR}{\mathbb{R}}
\newcommand{\EE}{\mathbb{E}}

\newcommand{\bigO}{\mathcal{O}}

\newcommand{\tr}[1]{\textrm{\normalfont tr}\left(#1 \right)}

\newcommand{\err}{\varepsilon}

\newcommand{\lambdamax}{\lambda_{\max}}
\newcommand{\lambdamin}{\lambda_{\min}}

\newcommand{\smooth}[1]{\widetilde{#1}}
\newcommand{\pred}[1]{\widehat{#1}}
\newcommand{\pes}{\mbox{{P-ES}}}

\newcommand{\githuburl}{www.github.com/estherrolf/p-es}

\newcommand{\figpath}{}
\graphicspath{}

\extrafloats{100}

\newtoggle{arxiv}
\toggletrue{arxiv}

\title{Post-Estimation Smoothing: \\ A Simple Baseline for Learning with Side Information}

\author{Esther Rolf\thanks{\texttt{esther\_rolf@berkeley.edu}} \\ UC Berkeley\and
Michael I. Jordan\\ UC Berkeley\and Benjamin Recht\\ UC Berkeley}

\begin{document}
\date{}

\maketitle

\begin{abstract}
  %!TEX root = ../aistats_draft/pes_aistats.tex
\vspace{-0.5em}
Observational data are often accompanied by natural structural indices, such as time stamps or geographic locations, which are meaningful to prediction tasks but are often discarded. 
We leverage semantically meaningful indexing data while ensuring robustness to potentially uninformative or misleading indices. 
We propose a post-estimation smoothing operator as a fast and effective method for incorporating structural index data into prediction.
Because the smoothing step is separate from the original predictor, it applies to a broad class of machine learning tasks, with no need to retrain models.
Our theoretical analysis details simple conditions under which post-estimation smoothing will improve accuracy over that of the original predictor. 
Our experiments on large scale spatial and temporal datasets highlight the speed and accuracy of post-estimation smoothing in practice. Together, these results illuminate a novel way to consider and incorporate the natural structure of index variables in machine learning. %tasks.
\end{abstract}

\section{INTRODUCTION}
% !TEX root=../pes_aistats.tex

The canonical machine learning setup models pairs of features and labels as originating from some underlying distribution,  $\{x_i,y_i\} \sim \mathcal{D} (x,y)$; the problem is to learn a predictor $\pred{y}(x)$ which describes $y$ as faithfully as possible. 
However, a recent narrative in machine learning is that well-annotated, large-scale datasets are rare, whereas less curated data are abundant; this has led to a taxonomy of supervision including \mbox{distant-,} weak-,  and semi- supervision. % \cite{zhou2017brief}. 
Whether labels are noisy by nature (distant) \citep{mintz2009distant}, programmatically generated (weak) \citep{ratner2016data}, or missing altogether (semi) \citep{zhu05survey}, 
it stands that characteristics of some data necessitate making use of additional sources of constraints.

Semi-supervised methods in particular aim to leverage unlabeled data to elicit an underlying structure which can aid  prediction \citep{singh2009unlabeled}. In practice, however, semi-supervised methods can be computationally expensive, and are sensitive to distribution shifts \citep{oliver2018realistic}. 
We propose to use readily-available data that is inherently structural, and apply a robust post-processing method which is independent of the original predictor to incorporate this structure.  

We consider scenarios where each datum $(x,y)$ has an associated index $t$ with some linking or semantic meaning. 
We thus represent observations as triplets: 
\begin{align*}
%\label{eq:setup}
\{x_i, y_i, t_i\} \quad i= 1,...,n
\end{align*}
%\vspace{-2mm}
Examples of such triplets include  \{image, annotation, frame number\} in video prediction, \{house attributes, price, address\} in  house price prediction, and \{document, sentiment, keywords\} in sentiment analysis. 
While intuition suggests that index variables $t$ may be \emph{correlated} with the label values $y$ and thus are highly informative to the prediction task, in many cases they are not well suited as \emph{predictors} of $y$ without major modification. For example, in object detection in videos, we may expect objects to move smoothly across frames, but the frame number itself does not carry predictive power from one video to another.

% Simply appending $t$ as column to a feature matrix could cause a predictor trained on both features and locations to over-rely on the location data, resulting in ``out-of-sample'' generalization issues for points which otherwise would have been "within sample." 
% As an illustrative example of potential over-reliance on $t$, imagine a video where a ball is tossed into the air, and the goal is to track the position of the ball over each frame. We could learn to estimate ball's position by fitting a parabola on $t$, the frame number of each image number of the video, or we could estimate the center of the ball from a photo at each frame. The former will fail as soon as we try to predict frames after the ball has hit the ground and begins bouncing, but the latter will likely learn a jittery estimation of location per frame. Simply throwing out the location indices $t$ is thus also undesirable. 

We aim to leverage the structural information encoded in $t$ without over-relying on it. This motivates a main question of our work: \emph{how can we utilize the dependence of $x$ and $y$ on $t$ even for predictors that might ignore or underestimate such dependence?}  We propose a \emph{post-estimation smoothing} (\pes) operator $S(t)$ that only depends on $t$ to obtain smoothed predictions:
\begin{align*}
\smooth{y} = S(t) \pred{y}(x).
%\label{eq:smooth_intro}
\end{align*}
Decoupling smoothing $S(t)$ from the initial feature-based prediction step $\pred{y}(x)$ allows us to efficiently smooth any off-the-shelf model. \pes\ applies to any precomputed predictions made over time or space, %or similar variables 
regardless of the original predictive model. The ease of applying \pes\ facilitates robust and reproducible incorporation of index variable structure in predictions.

%

%\subsection{Problem statement and contributions}
\paragraph{Problem Statement}
%!TEX root = ../aistats_draft/pes_aistats.tex
% Sec.~\ref{sec:setup} details the formal setup of the problem we analyze. Sections~\ref{sec:known_distributions} and~\ref{sec:bias_and_variance} characterize beneficial regimes of smoothing under the lens of data-generating distributions (for what types of data is smoothing helpful) and the lens of which properties of smoothers are helpful for large classes of data (where the data-generating distributions are probably not exact).

% \subsection{Setup}
% \label{sec:setup}
 
 \vspace{-0.5em}
Throughout this work we consider the setting in which we have a dataset indexed by $t_i \in \RR^l$, as well as predictions $\pred{y}_i \in \RR$ associated with each index.
It is natural to consider that there is also a set of features ${x_i} \in \RR^d$ and model ${f}: \RR^d \rightarrow \RR$ from which predictions $\pred{y} = {f}(x)$ were generated; %for reasons stated above 
we take this as given and work with directly with $\pred{y}$.

We study the post-prediction application of a \pes\ matrix operator $S(t): \RR^{n} \rightarrow \RR^{n}$ to form smoothed predictions, $\smooth{y} := S(t) \pred{y}$, such that the $\smooth{y}$ are closer to the true labels $y$ than the original unsmoothed predictions $\pred{y}$ are. 
In our theoretical analysis in Section~\ref{sec:analysis}, this is measured using the expected mean-squared error: 
$\EE\left[\tfrac{1}{n}\|\smooth{y} -y \|_2^2\right] = \EE\left[\tfrac{1}{n}\sum_{i=1}^n (\smooth{y}_i - y_i)^2\right]$,
while experiments in Section~\ref{sec:experiments} consider different accuracy metrics suitable to different contexts. 
%It will be illuminating to model observations, predictions, and prediction error residuals as stochastic processes indexed by variable $t$, so that 
%$\pred{y}(t) = y(t) + \err(t)$.

%Post-estimation smoothing is a powerful technique from which to study locality in data. 
Our main contributions are:
\iftoggle{arxiv}{}{\vspace{-1em}}
\begin{itemize}
% \item \pes, an efficient post-process smoothing operator for utilizing structural index variables, which is applicable to any predictor. 
% \item Theoretical results proving that under mild conditions, \pes\ will improve accuracy relative to the original predictor (\Cref{claim:smoothing_helps}), and characterizing when it can greatly increase predictive accuracy (\Cref{claim:unconstrained_opt}).

\item The formulation of a structural-index-based post-process smoothing procedure, \pes, which is applicable to any predictor. 
\item Theoretical results proving that under mild conditions, \pes\ will improve accuracy relative to the original predictor (\Cref{claim:smoothing_helps}), and characterizing when a linear smoothing operation can greatly increase predictive accuracy (\Cref{claim:unconstrained_opt}).
 \item Experiments on large-scale datasets for human pose estimation and house price prediction demonstrating that \pes\ improves accuracy of state-of-the-art predictors at minimal extra cost. 
\end{itemize}
%These contributions are made possible by the decoupling of the application of index variables as seperate from the original predictor, so that we utilize structural indexing variables for encoding local structure,. 
% make the point that because we only need structure to be on the variables we think there should be structure on, we reduce ability to oberfit.
%
These contributions are made possible by incorporating the general index variables separately from the feature based predictor. This results in a fast, accurate, and robust method for local variance reduction, with the potential to change how we consider and leverage structural variables in machine learning predictions.

More broadly, we demonstrate the effectiveness of a simple method that extends and generalizes previous scholarship in locality-based semi-supervised learning and nonparametric regression, applied in a modern context of abundant but weakly predictive data. Given recent exposition of the systematic underreporting of simple baselines \citep{dacrema2019we,mania2018simple}
 in machine learning and especially in semi-supervised learning~\citep{oliver2018realistic}, it is worth considering P-ES as a theoretically motivated and easily implementable baseline for semi-supervised learning and smoothing in large scale, real-data contexts. 
 %We evidence this in our pose estimation example (sec. 4.2), where our simple baseline significantly outperforms the provided (more complicated) baseline in the original paper.

 \section{RELATED WORK}
 \label{sec:related_work}
%!TEX root = ../aistats_draft/pes_aistats.tex
%A recent narrative in machine learning is that well-annotated, large scale datasets are rare, whereas less curated data are abundant; this has lead to a taxonomy of supervision including distant-, weak-,  and semi- supervision \cite{zhou2017brief}. 
%Some labels are noisy by nature (distant) \cite{mintz2009distant}, programatically generated (weak) \cite{ratner2016data}, or missing altogether (semi) \cite{zhu05survey}. These paradigms highlight that while data of different types are useful, the characteristics of some data necessitate clever crafting of predictive pipelines.

%To do so, 
Semi-supervised learning (SSL) methods leverage large amounts of unlabeled data along with some labeled data under a local consistency assumption: instances which are near to each other should have similar label values.
Distance is commonly determined with respect to an underlying manifold or graph defined by the features \citep{belkin2004semi,zhu2003semi}. 

To encourage local consistency of predictions,
\cite{belkin2006manifold} add a Laplacian regularization term to least squares and support vector machines, and \cite{jean2018semi} add a spatial regularization to the loss function of deep neural nets.
There is also a considerable amount of work incorporating additive consistency regularization and similar notions in neural nets \citep{tarvainen2017mean, bachman2014learning,grandvalet2005semi}. As noted by \cite{oliver2018realistic}, however, such methods can be sensitive to distribution shifts and require large validation sets and heavy computation to tune parameters; thus they are often poorly suited for ``real-world'' applications.

Unfortunately, adding a local consistency regularization term multiplies the number of parameter configurations in the optimization problem, and only works for predictors with an explicit objective function. For example, it  not straightforward to add a spatial consistency term to random forests. 
An alternative method, Gaussian harmonic energy minimization (HEM) \citep{zhu2003semi}, augments an underlying graph with noisy predictions, and solves for spatially consistent predictions on this larger graph. 
The local and global consistency (LCG) algorithm \citep{zhou2004learning} solves a similar optimization problem iteratively.

\citet{singh2009unlabeled} show that unlabeled data is useful in SSL precisely when it illuminates the underlying structure of the data beyond what was discernible by the labeled data alone.{}
However, if modeling assumptions incorrectly summarize the true structure of the data, unlabeled data can be misleading and even degrade performance \citep{cozman2002unlabeled, oliver2018realistic}. One approach to mitigate this is to fortify semi-supervised learning methods to be robust to this mismatch \citep{li2014towards}; we obtain robustness by decoupling feature-based prediction from a nonparametric incorporation of structural indices.
% Inspired by %this understanding of 
% the potential power of utilizing an underlying structure in data, and the responsibility of incorporating this structure faithfully to the true data, we propose a nonparametric smoothing step that separates data types that are characteristically structural from those that are predictive. 

The literature on nonparametric regression methods is extensive \citep{alexandretsybakov2010,gyorfi2006distribution}; we focus on two prominent approaches.
Gaussian Process Regression (GPR) 
%is a regression method which 
places a Gaussian prior on label covariances, specified by feature variables \citep{williams2006gaussian}. GPR has been widely adopted and extended in the geospatial statistics community, under the names of ``kriging'' and ``inverse distance interpolation'' \citep{babak2009statistical,lu2008adaptive}.
%, of which there are more sophisticated methods for heterogeneously spaced observations \cite{lu2008adaptive}.
As pointed out by \cite{banerjee2008gaussian}, when applied to large datasets, GPR has large computation and memory requirements, or necessitates approximations \citep{williams2006gaussian}.

Kernel smoothing \citep{alexandretsybakov2010} is another type of nonparametric regression in which predictions are locally weighted averages of observations. Of particular note is the Nadaraya-Watson estimator \citep{nadaraya1964estimating,watson1964smooth} in which weights are determined by a kernel relation on all instances. GPR, Laplacian regularized least squares \citep{belkin2006manifold}, HEM~\citep{zhu2003semi}, and exact LCG~\citep{zhou2004learning} can all be cast as instances of linear smoothing operators for which computing the smoothing matrix involves inverting a matrix of size $n \times n$.

% It is worth noting the similarity of our method to a single step of the iterative LCG method for classification~\citep{zhou2004learning} or ranking~\citep{zhou2004ranking}. This prior work allows us to connect our theoretical results in \Cref{claim:unconstrained_opt} with the properties of local and global consistency. Our method focuses on regression in a single-iteration smoothing operation, requiring new analysis techniques. To the best of our knowledge, no single-iteration, inverse-free method for SSL applicable to pre-made predictions has been proposed.

Lastly, recent work on the statistical optimality of data interpolation in machine learning \citep{belkin2018does} highlights that averaging methods are quite powerful for  prediction. We apply a locally-weighted average to \emph{predictions} themselves, which are the output of some prior model. 
Like semi-supervised learning methods, \pes\ encodes spatial consistency properties, but takes the perspective of refining given predictions with minimal restrictions on the underlying structure.
In light of the previous work, \pes\ can be seen as a fast and robust way to leverage structure, and to interpolate. 
Application-specific references are provided in Sec.~\ref{sec:experiments}.

\section{ANALYSIS}
\label{sec:analysis}
% !TEX root=../aistats_draft/pes_aistats.tex
Here we answer the questions  
\emph{(i) how should we form a useful post-estimation smoothing matrix while maintaining robustness to possible distributional misspecification?}
and
\emph{(ii) for what data distributions and predictors is linear smoothing beneficial?} 
Throughout the analysis, 
%it will be illuminating to
we model true values $y$, predictions $\pred{y}$, and error residuals $\err$ as stochastic processes indexed by $t$:
\begin{align}
\label{eq:gen_process}
\pred{y}(t) = y(t) + \err(t) ~.
\end{align}

\subsection{Accuracy Increases with General Smoothing Matrices}
	\label{sec:distribution_independent_bounds}
	% \inputtex{bias_variance}
While we may have strong intuition that there is \emph{some} locality-based structure in certain domains, the choice of distributional priors governing this structure will most often be inexact. 
We use a matrix $W(t) \in \RR^{n \times n}$, where weights $W_{ij}$ denote how much the $j^{th}$ prediction should contribute to a smoothed estimate for the $i^{th}$ instance, depending on the values of $t_i$ and $t_j$. 

\Cref{claim:smoothing_helps} below shows that using a reasonable 
weight matrix $W(t)$ which captures correlation in the underlying data can improve performance. A key insight is that shrinking $W$ towards the identity matrix tempers potential misspecification gracefully. Therefore, we form our smoothing matrix as the convex combination:
\begin{align}
\label{eq:S_c}
S_c(t) = c \cdot W(t) + (1-c) \cdot I ~, 
\end{align}
where in practice $c \in [0,1]$ can be chosen through cross-validation along with any parameters of $W$.%, and $W$ is a function of the index variables $t$. 

For any weight matrix, define the following quantities: $\gamma(\err,W)$, which describes the amount by which $W$ acts as a zero operator on the errors, and $\beta(\err,W; y)$, which describes the amount by which $W$ acts as the identity operator on the true labels, both scaled by $\EE[ \|\err\|_2^2]$:
% for convenience:
% \begin{align*}
% \gamma(\err,W) &:= \EE[\err^\top W \err]/\EE[ \|\err\|_2^2]
% \\
% \beta(\err,W) &: = {\EE[\err^\top (W-I)y ]}/{\EE\left[\|\err\|_2^2\right]}
% \end{align*}
\begin{align*}
\gamma(\err,W) &:= \EE[\err^\top W \err]/\EE[ \|\err\|_2^2],  \\
%\hspace{4em}
\beta(\err,W; y) &: = {\EE[\err^\top (W-I)y ]}/{\EE\left[\|\err\|_2^2\right]} ~. 
\end{align*}
Intuitively, we want to use a weight matrix $W$ such that both $\gamma$ and $\beta$ are small, so that $W$ averages out erroneous error signals while decreasing correlation between $y$ and $\err$. \Cref{claim:smoothing_helps} shows that an imperfect $W$ will suffice, so long as the sum $\gamma+ \beta$ is controlled.

\begin{restatable}[]{thm}{smoothinghelps}
\label{claim:smoothing_helps}
Given any predictor $\pred{y}$ of $y$ with error residuals
satisfying $\EE\left[ \|\err \|_2^2\right] \neq 0$, 
and any weight matrix $W$ %with eigenvalues at least 0 and 
satisfying 
$\gamma(\err,W)+ \beta(\err,W; y)< 1$,
%(i) $\gamma(\err,W) < 1$, and
%(ii) $\beta(\err,W) < 1 - \gamma(\err,W)$
% \begin{enumerate}[(i)]
% \item $\gamma(\err,W):= \EE[\err^\top W \err]/\EE[ \|\err\|_2^2] < 1$, and
% \item $\beta(\err,W) : = {\EE[\err^\top (W-I)y ]}/{\EE\left[\|\err\|_s^2\right]} < 1 - \gamma(\err,W)$ 
% \end{enumerate}
there exists a constant $c \in (0,1]$ such that the smoothing matrix
$
S_c = c\cdot W + (1-c)\cdot I
$
strictly reduces expected MSE:
\begin{align*}
\EE\left[\tfrac{1}{n}\|S_c \pred{y} - y\|_2^2\right] < \EE\left[\tfrac{1}{n}\|\pred{y} - y\|_2^2 \right]~.
\end{align*}
\end{restatable}
\begin{proof}[(Proof sketch for $\beta = 0$.) ] 
For unbiased estimators $\pred{y}$ with errors $\err$ that are independent of the labels, $\beta = 0$ and the objective decomposes as  %(recall $\EE[\| \pred{y}-y\|_2^2 ] = \EE[\|\err\|_2^2 ]$):
\begin{align*}
&\EE\left[\|S_c \pred{y} - y\|_2^2 - \| \pred{y} - y\|_2^2\right]  \\
 %&= \|c (W\hat{y} - y) + (1-c)(\err)\|_2^2 \\
%&= 
%c^2 \EE\left[\|W\pred{y} - y\|_2^2\right] + \left((1-c)^2 - 2\gamma c(c-1) - 1\right)\EE\left[\|\err\|_2^2 \right] \\
&\quad  \leq c^2\left(\EE\left[\|W\pred{y} - y\|_2^2\right] + 2(1-\gamma )\EE\left[\|\err\|_2^2 \right] \right)
 \\
& \hspace{4em} 
+  
2c(\gamma-1)\EE\left[\|\err\|_2^2 \right] ~.
% c^2\left(\EE\left[\|W\hat{y} - y\|_2^2\right] + (1-\gamma )\EE\left[\|\err\|_2^2 \right] \right)+ 
% 2c(\gamma-1)\EE\left[\|\err\|_2^2 \right]
\end{align*}
% \end{align*}
% Recall $\gamma := \EE\left[(\err^\top W\err)\right]/\EE\left[(\err^\top \err)\right] $.
% Then $0 \leq \lambdamin(W) \leq \gamma \leq \lambdamax(W) \leq 1$ and
% \begin{align*}
% \EE\left[\|S_c \hat{y} - y\|_2^2\right] - \EE\left[\|\err\|_2^2 \right]
% &\leq
% c^2\left(\EE\left[\|W\hat{y} - y\|_2^2\right] + (1-\gamma )\EE\left[\|\err\|_2^2 \right] \right)+ 
% 2c(\gamma-1)\EE\left[\|\err\|_2^2 \right]
% \end{align*}
The upper bound is a convex quadratic function in $c$ with 
optimum at 
\begin{align*}
c^*  = \frac{(1-\gamma)\EE\left[\|\err\|_2^2 \right]}{\left(\EE\left[\|W\pred{y} - y\|_2^2\right] + 2(1-\gamma )\EE\left[\|\err\|_2^2 \right] \right)} 
\end{align*}
By the theorem conditions, $\gamma < 1$, so that $c^* \in (0,1]$.
The resulting upper bound is then given by
\begin{align*}
\hspace{-1mm}
&\EE\left[\tfrac{1}{n}\|S_{c^*} \pred{y} - y\|_2^2 - \tfrac{1}{n}\|\pred{y}-{y}\|_2^2 \right] \\
&\leq 
% c^2\left(\EE\left[\|W\hat{y} - y\|_2^2\right] + (1-\gamma )\EE\left[\|\err\|_2^2 \right] \right)+ 
% 2c(\gamma-1)\EE\left[\|\err\|_2^2 \right]
% \\
% % &= \frac{(1-\gamma)^2\EE\left[\|\err\|_2^2 \right]^2 - 2(1-\gamma)^2 \EE\left[\|\err\|_2^2 \right]^2}{\left(\EE\left[\|W\hat{y} - y\|_2^2\right] + (1-\gamma )\EE\left[\|\err\|_2^2 \right] \right)}\\
% &= 
% -\frac{(1-\gamma)^2\EE\left[\|\err\|_2^2 \right]^2}{\left(\EE\left[\|W\hat{y} - y\|_2^2\right] + (1-\gamma )\EE\left[\|\err\|_2^2 \right] \right)}
-
\frac{(1-\gamma)^2\EE\left[\|\err\|_2^2 \right]^2}
{n (\EE\left[\|W\pred{y} - y\|_2^2\right] + 2(1-\gamma )\EE\left[\|\err\|_2^2 \right])} 
< 0~.\qedhere
\end{align*}

\end{proof}
We present the full proof in Appendix~\ref{sec:proof_of_thm1}.

\Cref{claim:smoothing_helps} inverts the standard statistical smoothing analysis \citep{simonoff1998smoothing,alexandretsybakov2010}---which, provided generative processes as in Eq.~\eqref{eq:gen_process}, calculates the bias and variance of the resulting smoothed estimator---and instead characterizes properties of the underlying signal $y(t)$, the prediction errors $\err(t)$, and the weight matrix $W$ that make smoothing beneficial. 
As in standard kernel smoothing regression \citep{wand1994kernel}, in \pes\ we are willing to tolerate an increase in the bias of our predictor, so long as the variance decreases.
Formulating the conditions in terms of $\gamma$ and $\beta$ allows us to assess this trade-off in terms of the conditions on the prediction errors directly.

We can guarantee that $\gamma(\err,W) \leq 1$ by ensuring $\lambdamax(W) \leq 1$, for example by taking any right-stochastic matrix (here $\lambdamax(\cdot)$ denotes the maximum eigenvalue).  Controlling $\beta(\err, W; y)$ %is in general harder as it 
depends on both the true values and the errors in predictions. A sufficient 
%(but not necessary) 
condition to achieving $\beta \leq c$ is to pick $W$ which satisfies $\EE\left[ \| (W-I)y \|_2^2\right] \leq c^2\EE \left[ \| \err \|_2^2\right]$, encoding that we tolerate a deviation in labels due to $W$ limited by the magnitude of errors that can potentially be reduced.  
See Appendix~\ref{sec:gamma_beta_conditons} for ways to ensure $\gamma + \beta <1$.

In general, the condition $\beta + \gamma < 1$ represents a trade-off in choosing a weight matrix that acts approximately as a zero matrix with respect to the errors (small $\gamma$), while acting close to an identity matrix with respect to the true values (small $\beta$). In order to keep the sum small, $W$ needs to incorporate knowledge in the structure of the domain-specific labels, $y$, as well as the distribution of prediction errors, $\err$, for a given predictor. For all experiments (Section~\ref{sec:experiments}),  we use the Nadaraya-Watson smoothing matrix with a Gaussian kernel (Eq.~\eqref{eq:S_NW}). This matrix is right-stochastic, and the Gaussian kernel encodes the constraint that nearby data points (measured with respect to structural index variable $t$) should have similar label values $y$.

The best-case reduction in MSE attainable by \pes\ is bounded in the final line of the proof of \Cref{claim:smoothing_helps} (for $\beta \neq 0$, see Appendix~A.1). The MSE reduction depends on covariances between $W$, $y$, and the error residuals in $\hat{y}$. This motivates us to study the form of the optimal smoothing operator and the resulting expected error reduction, when these covariances are known.

\subsection{Optimal P-ES for Known Distributions}
	 \label{sec:known_distributions}
	% \inputtex{linear_analysis}
Having proposed a \pes\ matrix $S_c$ in Section~\ref{sec:distribution_independent_bounds}, we now study the form of an optimal linear smoothing matrix $S^*$ when the distributions governing the labels and error residuals are known. This reinforces the high-level structures we wish to capture in $S_c$,
and provides a baseline for simulation experiments in Section~\ref{sec:simulations}. Denote the cross-correlation matrices $K$ element-wise as 
$ K_{xy}[t,s] = \EE\left[ x(t) y(s)\right]$.
% and provides a baseline for comparing our chosen $S$ to the best-case performance in simulated experiments (Sec.~\ref{sec:simulations}).
\begin{restatable}%[Optimal linear smoother]
{lem}{unconstrainedopt}
\label{claim:unconstrained_opt}
For a predictor $\pred{y}$ of $y$ with error residuals distributed as \mbox{$\err(t) = \pred{y}(t)-y(t)$}, when $K_{\pred{y}\pred{y}} \succ 0$, the optimal linear smoothing matrix has the form 
\begin{align}
S^* &= \argmin_{S \in \RR^{n \times n}} \EE\left[\tfrac{1}{n}\| S\pred{y} - y \|_2^2\right] \nonumber\\
 %& = K_{y \pred{y}} (K_{\pred{y}\pred{y}})^{-1 } 
 &= I - (K_{\err\err} +K_{y \err})^\top(K_{y y} +K_{y\err} + K_{\err y} + K_{\err\err})^{-1 } ~.
 \nonumber
\end{align}
The expected MSE reduction of applying $S^*$ versus using the original predictions $\pred{y}$ is always non-negative, and is given by
\begin{align*}
\tfrac{1}{n}
 \EE\left[
 \|\pred{y} - y \|_2^2] - \| S^* \pred{y} - y \|_2^2\right] 
 = 
 %\tfrac{1}{n}\tr{(K_{\err\err} + K_{y\err })^\top (K_{yy} + K_{\err\err} + K_{\err y} + K_{y \err})^{-1}(K_{\err\err} + K_{y\err })}  \\ %\\
  \tfrac{1}{n} \tr{K_{\pred{y}y}^\top (K_{\pred{y}\pred{y}})^{-1}K_{\pred{y}y}} ~. \nonumber
 %&\geq \frac{\tr{(K_{\err\err} + K_{y\err })^\top(K_{\err\err} + K_{ y \err})}}{ n \cdot \tr{K_{yy} + K_{\err\err} + K_{\err y} + K_{y\err }}}
 \end{align*}
%\end{lem}
\end{restatable}
% \begin{proof}
% We prove \Cref{claim:unconstrained_opt} in Appendix~\ref{sec:proof_of_lem1}.
% \end{proof}
We present the proof of \Cref{claim:unconstrained_opt} in Appendix~\ref{sec:proof_of_lem1}. 

\Cref{claim:unconstrained_opt} shows that smoothing can reduce prediction error associated with $K_{\err\err}$ and  $K_{y\err}$, but that the extent to which errors can be smoothed out depends on the forms of $K_{yy}$ and $K_{y \err}$. 
% To see this, take $K_{\err\err} = \sigma^2_{\err}I$, $K_{\err y} = 0$, then the optimal smoothing matrix is $K_{yy} (K_{yy} + \sigma^2_{\err}I)^{-1}$, and the reduction is 
% \begin{align*}
% \EE\left[\frac{1}{n}\|\pred{y} - y \|_2^2 - \frac{1}{n}\| W^* \pred{y} - y \|_2^2\right] 
% %&= \frac{1}{n}\tr{\sigma^4_{\err}(K_{yy} + \sigma^2_{\err}I)^{-1}} \\
% &= \frac{\sigma^4_{\err}}{n} \cdot \tr{(K_{yy} + \sigma^2_{\err} I)^{-1}} 
% %& =\sigma_{\err}^2 \cdot \sum_{i=1}^n \frac{1}{\lambda_i(K_{yy}) + \sigma^2_{\err} } \\
% &\geq \sigma_{\err}^2\cdot\left( \frac{\sigma_{\err}^2}{\sigma^2_{\err} + \lambda_{\max}(K_{yy})}\right)
% \end{align*}
The following example underscores this point for an illustrative data generating model and sets the stage for simulation experiments in Section~\ref{sec:simulations}. The main details of the example are given here, with more extensive exposition in Appendix~\ref{sec:appendix_example}. 
\begin{exmp}
\label{exmp:linear}
Consider zero-mean stochastic processes $x(t)$ and $y(t)$ which are dependent on a third zero-mean hidden process $z(t)$, but with independent additive Gaussian noise. In particular:
% \begin{align}
% \label{eq:linear_example}
%  z &\sim \mathcal{N}(0, K_{zz}) \\
%  x(t) &= z(t) + \omega(t), \quad \omega(t) \sim_{i.i.d.} \mathcal{N}(0,\sigma^2_x) \nonumber \\
%  y (t) &= c \cdot z(t) + \mu(t), \quad \mu(t) \sim_{i.i.d.} \mathcal{N}(0,\sigma^2_y) \nonumber
% \end{align}
\begin{align}
\label{eq:linear_example}
z &\sim \mathcal{N}(0, K_{zz})& \\
 x(t) &= z(t) + \omega(t), \quad &\omega(t) \sim_{i.i.d.} \mathcal{N}(0,\sigma^2_x) \nonumber\\
 y (t) &= c \cdot z(t) + \mu(t), \quad  &\mu(t) \sim_{i.i.d.} \mathcal{N}(0,\sigma^2_y) ~. \nonumber 
\end{align}
The autocorrelation matrices show that there is shared variation due to the ``hidden" process $z$:
% \begin{align*}
% K_{xx}[t,s] &= K_{zz}[t,s] + K_{\omega \omega}[t,s], &
% K_{yy}[t,s] &= c^2 K_{zz}[t,s] + K_{\mu \mu }[t,s],
% &
% K_{xy}[t,s] &= c K_{zz}[t,s] 
% \end{align*}
\begin{align*}
K_{xx} &= K_{zz} + K_{\omega \omega}, &
K_{yy} &= c^2 K_{zz} + K_{\mu \mu },
&
K_{xy} &= c K_{zz} ~.
\end{align*}
Without any specific knowledge of the covariance structure in $z$, this could be modeled with an ``errors in variables'' model, for which total least squares (TLS) gives a statistically consistent estimator of $c$. In the appendix, we show that as $n$ grows large, the expected MSE of the TLS predictions approaches 
\begin{align*}
\EE\left[\tfrac{1}{n}\|\pred{y}_{TLS} - y\|_2^2\right]
&\approx  \sigma^2_y + c^2 \sigma^2_x ~,
\end{align*}
whereas invoking~\Cref{claim:unconstrained_opt}, the expected smoothed performance using $S^*$ approaches \begin{align*}
&\EE\left[\tfrac{1}{n}\|S^*\pred{y}_{TLS} - y\|_2^2\right]
\\ 
& ~~~~
\approx \sigma^2_y + c^2\sigma_x^2  \left(1 
- \tfrac{1}{n}  \tr{(\sigma_x^{-2}K_{zz} +I )^{-1} } \right) \geq \sigma_y^2~.
\end{align*}
 Using the first line above, the expected MSE reduction is approximately
 $\frac{c^2\sigma_x^2}{n} \tr{\left(\sigma_x^{-2}K_{zz} +I \right)^{-1} }$ which is strictly positive for $\sigma_x^2 > 0$, and increasing with $\sigma_x^2$. 

%The o.l.s. estimate is subject to attenuation bias, and thus is more difficult to form $W^*$ (see Appendix Section~\ref{sec:appendix_example} for details on using the o.l.s. estimator in this setting.)
\end{exmp}

% \begin{remark}%[A suitable approximation to the optimal smoothing matrix]

% In practice we use the Nadaraya-Watson (NW) estimator as $\hat{W}$; this is equivalent to approximating $(K_{yy} + K_{\err\err})^{-1}$ as a diagonal matrix with entries equal to the row sums of $K_{yy}$.
% % \begin{align*}
% % \hat{W} = K_{yy} \diag{K_{yy}\vec{1}}^{-1}
% % \end{align*}
% This $\hat{W}$ is a right-stochastic matrix so that elements of $\hat{W}y$ are locally weighted averages of elements of $y$, and $\lambdamax(\hat{W}) =1$. %The finite sample NW estimator is consistent, with second order convergence properties tending to 0 as $n \rightarrow \infty$ \cite{alexandretsybakov2010}.
% % \todo{reword}
% \end{remark}

\iftoggle{arxiv}{
  \begin{figure*}[!h] %  figure placement: here, top, bottom, or page
   \centering
   \includegraphics[width=.95\textwidth]{\figpath 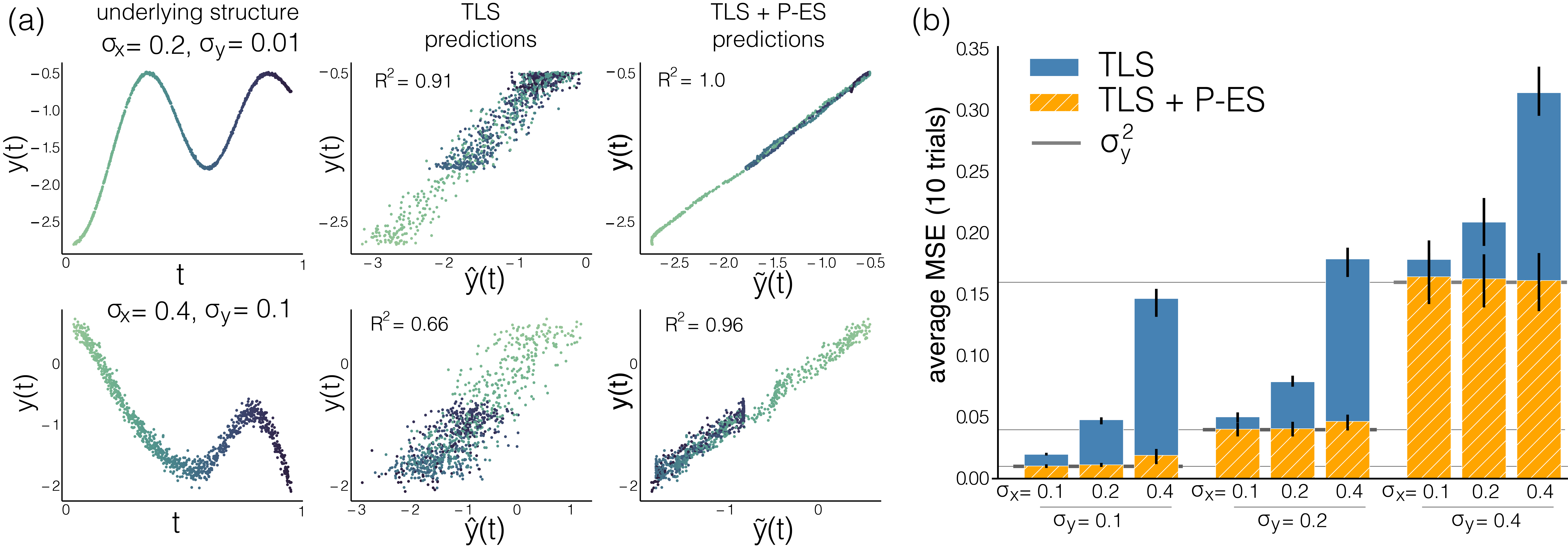}
      \caption{Simulation results. (a) 
      Two examples of structure in $z$ (left column), where a TLS estimator recovers structure (middle column), but is  improved upon using \pes\ (right column). (b) Aggregate performance over different noise parameters for unsmoothed and \pes\ estimates, compared to a lower bound of $\sigma_y^2$ for any linear smoother. Vertical black lines show min and max over 10 trials.
      }
   \label{fig:simulation_panel_tls}
   \vspace{-.65em}
\end{figure*}
}
{
\begin{figure*}[!h] %  figure placement: here, top, bottom, or page
   \centering
   \includegraphics[width=.95\textwidth]{figures/sim_by_param_panel_tls.pdf}
      \caption{Simulation results. (a) 
      Two examples of structure in $z$ (left column), where a TLS estimator recovers structure (middle column), but is  improved upon using \pes\ (right column). (b) Aggregate performance over different noise parameters for unsmoothed and \pes\ estimates, compared to a lower bound of $\sigma_y^2$ for any linear smoother. Vertical black lines show min and max over 10 trials.
      }
   \label{fig:simulation_panel_tls}
   \vspace{-.65em}
\end{figure*}
}

\section{EXPERIMENTS}
  \label{sec:experiments}
  % !TEX root=../aistats_draft/pes_aistats.tex

We first use simulated experiments to study situations in which smoothing is beneficial and to demonstrate that a simple instantiation of Eq.~\eqref{eq:S_c} achieves close to optimal accuracy in these settings. We then apply \pes\ to predictions on real-world datasets with temporal and spatial structure: human-pose prediction in video (Sec.~\ref{sec:video_example}) and house-price prediction over space (Sec.~\ref{sec:housing_example}). \pes\ improves performance of all predictors we consider, including some that already incorporate locality. 
\pes\ compares favorably to statistical smoothing and SSL methods, both in predictive accuracy and computation time.
As a simple local-averaging weight matrix, all experiments use as $W$ the Nadaraya-Watson smoothing matrix with squared exponential kernel on $t$, where  $D_{ij}(t; \sigma) = e^{-\frac{1}{2\sigma^2} \| t_i - t_j\|_2^2}$:
%\vspace{-.5mm}
\begin{align}
\label{eq:S_NW}
 S_c(t;\sigma) &= c \cdot \textrm{diag}^{-1}\left(D\vec{1}\right) D + (1-c) \cdot I ~.
% \nonumber
\end{align}

  \subsection{Simulations}
  \vspace{-.8mm}
    \label{sec:simulations}
    % !TEX root=../aistats_draft/pes_aistats.tex

We return to the distribution defined in Eq.~\eqref{eq:linear_example} in Example~\ref{exmp:linear}, where the processes $x(t)$ and $y(t)$ are influenced by a third `hidden' process $z(t)$. We now make a specific assumption for the covariance of $z$: 
\begin{align*}
\vec{z} &= \mathcal{N}\left(\vec{0}, \Sigma(t)\right), \quad \Sigma_{ij}(t) = e^{- \frac{1}{2\sigma_z^2} (t_i-t_j)^2}  ~. 
\end{align*}

We take $t= [0,1/n, 2/n, \dots, (n-2)/n, (n-1)/n]$ with $n=2000$, and $\sigma_z=0.2$. Half of the points are chosen at random to form a training set from which we learn the total least squares (TLS) estimator $\pred{y}(x)$. The remaining $1000$ points are used to evaluate performance with and without \pes. To show the expressiveness of the matrix $S_c$, in simulations we pick the parameters $c,\sigma$ of $S_c$ so as to maximize performance on the evaluation set. In  Sections~\ref{sec:video_example} and ~\ref{sec:housing_example}, we pick parameters on a validation set before applying to a holdout set.

Figure~\ref{fig:simulation_panel_tls}(a) shows the process of \pes\ as local variance reduction. Each row shows a different setting of $\sigma_x, \sigma_y$. 
The leftmost column shows  observed labels $y(t)$ as a function of indices $t$. The middle and right columns show the TLS predictions, without and with \pes, respectively.  Errors in $\pred{y}$ that are made in the horizontal axis are reducible by smoothing, as $S_c(t)$ gives more weight to pairs closer in $t$ (similar hue in Figure~\ref{fig:simulation_panel_tls}(a)). 
The smoothed predictions $\smooth{y}$ exhibit a similar structure to the original predictions, with significantly reduced horizontal error bands.
The difference in the performance of the TLS estimator with and without \pes\ (Figure~\ref{fig:simulation_panel_tls}(b)) indicates that \pes\ reduces prediction errors that are uncorrelated with the index variable $t$.

% Fig.~\ref{fig:simulation_panel_tls}(B) shows aggregate performance over many values of 
% $\sigma_x$ and $\sigma_y$.
% Results match the analysis in Example~\ref{exmp:linear}; \pes\ reduces prediction errors associated with the random noise in $x$, but is limited by the observation noise in $y$. This is evidenced by large reductions in MSE for smoothed vs. unsmoothed predictors as $\sigma_x$ increases, with stable performance in smoothed predictions across $\sigma_y$ values. The closeness of the smoothed performance using the matrix $S_c$ from Eq.~\eqref{eq:S_NW} to the lower bounds $\sigma_y^2$ shows that the approximation is indeed suitable for this problem setting.

  \subsection{Human Pose Prediction in Video}
  \label{sec:video_example}
    % !TEX root=../pes_arxiv.tex

\iftoggle{arxiv}{
  \begin{table*}[t!]
  \caption{Holdout set performance for human pose estimation in video. Arrows indicate the direction of desired performance; bold numbers indicate the best performance for each metric. For all metrics, \pes\ predictions (italicized methods) have the best performance of all methods considered. Other methods are attributed as ``temporal, temporal + dynamics:'' \cite{kanazawa2019humanDynamics}, ``per-frame:'' \cite{kanazawa2019humanDynamics,kanazawa2018end}.
}
 \label{table:video_performance_test}
  \centering
  \begin{tabular}{lllllll}
    \toprule
 \multicolumn{1}{c}{}  & \multicolumn{4}{c}{3DPW} & \multicolumn{2}{c}{Penn Action}\\
     \cmidrule(r){2-5} \cmidrule(r){6-7}
  \multicolumn{1}{c}{method} &PCK $\uparrow$ & MPJPE $\downarrow$ & PA-MPJPE $\downarrow$ & Acc. Err. $\downarrow$ & PCK $\uparrow$ & Accel $\downarrow$ \\
    \cmidrule(r){0-0}  \cmidrule(r){2-5} \cmidrule(r){6-7}
per-frame & 84.06 & 129.95 & 76.68 & 37.41  & 73.17 & 79.91 \\
\textit{per-frame, with \pes} \  & 84.46 & 128.44 & \bf{75.84} & 20.46  & 73.74 & 48.22  \\
%temporal only & 82.59 & 139.19 & 78.35 & 15.15  & 71.16 & 29.30 \\ 
\hline
temporal & 82.59 & 139.19 & 78.35 & 15.15 & 71.16 & 29.30 \\
temporal + dynamics
& 86.37 & 127.08 & 80.05 & 16.42  & 77.88 & 29.66 \\ 
\textit{temp. + dyn. with \pes }\ & \bf{86.57} & \bf{126.14} & 79.73 & \bf{8.14}  & \bf{78.07} & \bf{4.96} \\    
\bottomrule
  \end{tabular}
\end{table*}
}
{
\begin{table*}[t!]
  \caption{Holdout set performance for human pose estimation in video. Arrows indicate the direction of desired performance; bold numbers indicate the best performance for each metric. For all metrics, \pes\ predictions (italicized methods) have the best performance of all methods considered. Other methods are attributed as ``temporal, temporal + dynamics:'' \cite{kanazawa2019humanDynamics}, ``per-frame:'' \cite{kanazawa2019humanDynamics,kanazawa2018end}.
}
 \label{table:video_performance_test}
  \centering
  \begin{tabular}{lllllll}
    \toprule
 \multicolumn{1}{c}{}  & \multicolumn{4}{c}{3DPW} & \multicolumn{2}{c}{Penn Action}\\
     \cmidrule(r){2-5} \cmidrule(r){6-7}
  \multicolumn{1}{c}{method} &PCK $\uparrow$ & MPJPE $\downarrow$ & PA-MPJPE $\downarrow$ & Acc. Err. $\downarrow$ & PCK $\uparrow$ & Accel $\downarrow$ \\
    \cmidrule(r){0-0}  \cmidrule(r){2-5} \cmidrule(r){6-7}
per-frame & 84.06 & 129.95 & 76.68 & 37.41  & 73.17 & 79.91 \\
\textit{per-frame, with \pes} \  & 84.46 & 128.44 & \bf{75.84} & 20.46  & 73.74 & 48.22  \\
%temporal only & 82.59 & 139.19 & 78.35 & 15.15  & 71.16 & 29.30 \\ 
\hline
temporal & 82.59 & 139.19 & 78.35 & 15.15 & 71.16 & 29.30 \\
temporal + dynamics
& 86.37 & 127.08 & 80.05 & 16.42  & 77.88 & 29.66 \\ 
\textit{temporal + dynamics with \pes }\ & \bf{86.57} & \bf{126.14} & 79.73 & \bf{8.14}  & \bf{78.07} & \bf{4.96} \\    
\bottomrule
  \end{tabular}
\end{table*}
}

Recent work has shown that improvements in human pose estimation \citep{zhang2019phd, kanazawa2019humanDynamics,dabral2018learning} and object detection and classification in videos \citep{prest2012learning,yucer2015reconstruction,zhu2017deep} can be obtained by encoding temporal consistency as part of a larger predictive pipeline. The intuition is that exploiting continuity of motion over video frames can reduce the noise in per-frame predictions. For example, a recent state-of-the-art method for pose estimation \citep{kanazawa2019humanDynamics} learns both a temporal encoder and temporal human dynamics as part of the predictive pipeline.  In the following experiment, we apply \pes\ to predictions from this model
as well as to predictions from a per-frame baseline model \citep{kanazawa2018end}.

We use the same validation and holdout splits for the 3D Poses in the Wild (3DPW) \citep{vonMarcard2018} and the Penn Action datasets \citep{zhang2013actemes} as in \cite{kanazawa2019humanDynamics}.
Before testing results on the holdout set,  smoothing parameters were chosen from
 $\sigma \in [0.5,1,2,3,4]$ frames, and $c \in [0.0,0.2,0.4,0.5,0.6,0.7,0.8,0.9,1.0]$ to maximize average validation accuracy measured by the key-point accuracy (PCK) metric  (see Appendix~\ref{sec:appendix_experiment_details} for details). 
%We chose to optimize this metric specifically because it has the least ``smooth'' interpretation of the four metrics. 

Holdout test set performance %for all metrics reported in \cite{kanazawa2018learning} 
is given in Table~\ref{table:video_performance_test}. PCK is an accuracy metric on key-points, MPJPE and PA-MPJPE measure error over predicted pose joints, and acceleration error penalizes high acceleration predictions (see Appendix~\ref{sec:appendix_experiment_details} for a discussion of the metrics). 
%We see that (as hypothesized in \cite{kanazawa2018learning}), 
%. 
While we optimized according to PCK, \pes\ improves performance in both models, across all metrics.

Smoothing confers greater gains in the time-agnostic per-frame model than the temporal dynamics model. 
Interestingly, the ``temporal model'' without human dynamics 
does worse in almost all metrics than the ``per-frame'' model that ignores frame number. 
%Had \pes\ been compared to in the original work, the conclusion regarding temporal smoothers without dynamics would have been different. 
This underscores our motivation, that temporal information must be encoded with care, as well as our claim that \pes\ is a suitable baseline for such tasks.

The optimal hyperparameter pairs chosen are given in Table~\ref{table:video_params_chosen}. 
%We allowed for different hyperparameter configurations between datasets to allow for data-dependent qualities like frame rate or average distance from subject. 
For both models, the optimal $\sigma$ was around $2$ frames, but the optimal $c$ for the per-frame predictions was much smaller for the per-frame model (avg. 0.45) than for the  model that already incorporated temporal structure (avg. 0.9). This may be because predictions for the temporal model are smoother, so that we do not alter the signal in predictions as much with \pes. 

\begin{table}[h]
  \caption{Optimal hyperparameter values on validation set per predictor for both datasets.}
   \label{table:video_params_chosen}
  \centering
  \begin{tabular}{cllll}
    \toprule
 \multicolumn{1}{c}{}  & \multicolumn{2}{c}{per-frame} & \multicolumn{2}{c}{temporal + dynamics }\\
    \cmidrule(r){2-3} \cmidrule(r){4-5}
     & 3DPW     & PA    & 3DPW     & Penn Action \\
    \cmidrule(r){0-0}  \cmidrule(r){2-5}
    $\sigma$ & 2     &  3     &  2     & 2     \\
    $c$     &0.5     & 0.4     & 0.8    & 1.0    \\
    \bottomrule
  \end{tabular}
  \iftoggle{arxiv}{}{\vspace{-.5em}}
\end{table}

In summary, \pes\ confers performance gains to both the per-frame model and the more accurate temporal and dynamics model, showing that \pes\ can improve performance even when the base estimator is a complex model incorporating locality.  
 % todo: explain our smoothing method applied to videos

  \subsection{Predicting House Price from Attributes}
  \label{sec:housing_example}
    % !TEX root=../pes_arxiv.tex

\begin{table*}[h]
  \caption{Comparison with nonparametric and semi-supervised methods, for 10 random trials with train, validation, and holdout sets of size $n=10,000$.
  % on predicting house price in ZTRAX data. 
  %For fairness and to ensure termination within a reasonable time frame, 
}
   \label{table:zillow_compared_to_ssl}
  \iftoggle{arxiv}{}{\vspace{-1.2em}}
  \begin{center}
  \begin{tabular}{lcll }
    \toprule
   &  model function ($f$) or & average holdout accuracy & average runtime \\
    method & post-processing ($pp$) form &mean (std) in $r^2 $& mean (std) in secs\\
   \cmidrule(r){0-0}  \cmidrule(r){2-2} \cmidrule(r){3-4}
kernel smoothing & $f(t,y)$ & 0.277 (0.185)& 20.6 (0.2) \\
GPR (Kriging) & $f(t,y)$  & 0.386 (0.011)& 1336.1 (4.6) \\
LapRLS \citep{belkin2006manifold} & $f(x,t,y)$ & 0.452 (0.012)& 1683.6 (12.4) \\
\hline
XGB & $f(x,y)$ & 0.458 (0.014)& 7.8 (0.1) \\
XGB + shrinkage & $pp(t,\pred{y}_{\textrm{\tiny{XGB}}})$ & 0.457 (0.014)& ~ + 0.0 (0.0) \\
\textit{XGB + \pes} & $pp(t,y,\pred{y}_{\textrm{\tiny{XGB}}})$ & 0.526 (0.015)& ~ + 27.0 (0.2) \\
HEM \citep{zhu2003semi} & $f(t,y, \pred{y}_{\textrm{\tiny{XGB}}})$ & 0.544 (0.015)& ~ + 898.6 (4.8) \\
\textit{HEM \citep{zhu2003semi} + \pes}& $pp \left(t,y, f(t,y, \pred{y}_{\textrm{\tiny{XGB}}}) \right)$ & 0.546 (0.015)& ~  ~ ~+ 287.8 (8.0) \\   
    \bottomrule
  \end{tabular}
    \end{center}
\end{table*}

\begin{figure}[h] %  figure placement: here, top, bottom, or page
   \centering
   \includegraphics[width=0.5\columnwidth]{\figpath 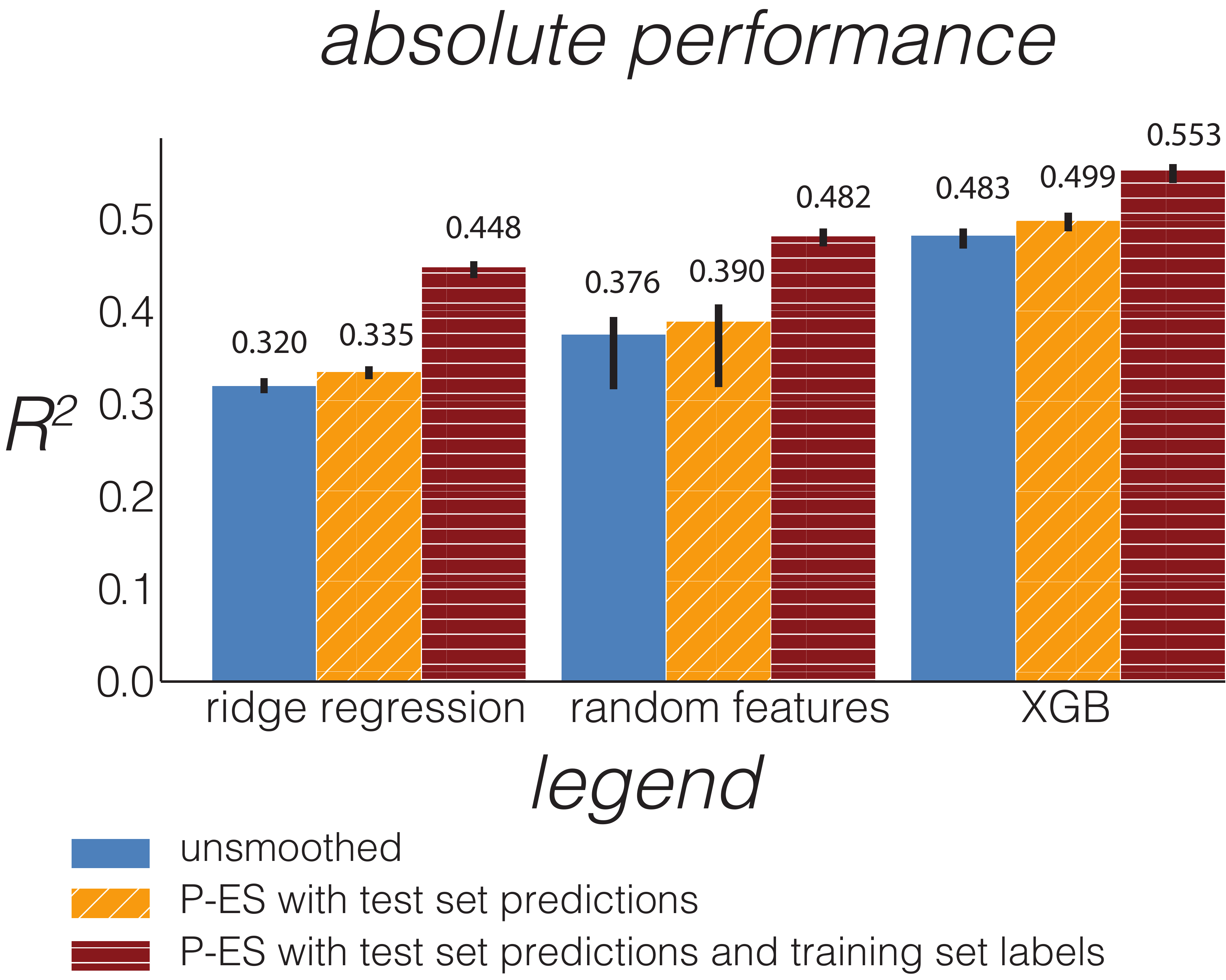}
      \caption{$\pes$ performance for various base predictors. 
      %\pes\ improves accuracy of all predictors. 
      %Train, validation, and test sets are of size is 20k; 
      Bars denote average performance; vertical black lines show min and max over 10 trials. All six average relative differences (unsmoothed - smoothed) are positive with p-value $< 1e^{-4}$, those that include the training set have p-value $<1e^{-7}$.}
      \label{fig:table_1_barchart}
      \iftoggle{arxiv}{}{\vspace{-1em}}
\end{figure}

The usefulness of applying semi-parametric techniques merging feature-based prediction and spatial regularization in predicting house prices has been documented from many perspectives \citep[see, e.g.,][]{caplin2008machine,dubin1998predicting,clapp2002predicting,can1992specification}. 
This motivates house price prediction as a domain in which to compare the performance of \pes\ and other methods exploiting spatial consistency.
%The task we consider is that of house price prediction, 
Using data on house sales from the Zillow Transaction and Assessment Database (ZTRAX) \citep{ztrax},
we first demonstrate the effectiveness of \pes\ on various machine learning regression methods (Figure~\ref{fig:table_1_barchart}), and then in comparison to standard semi-supervised learning techniques (Table~\ref{table:zillow_compared_to_ssl}). 

In this experiment we predict sale prices $y$ of single family homes, given features $x$ about the homes (e.g., number of bedrooms, year of home sale, etc.). Location $t$ is the latitude and longitude of the homes. 
After preprocessing (see Appendix~\ref{sec:appendix_experiment_details}), the dataset contains roughly 600,000 home sales at unique locations.
We test three diverse machine learning models: ridge regression, random feature regression~\citep{rahimi2009weighted}, and gradient boosted decision trees (XGB)~\citep{chen2016xgboost}, with and without post-estimation smoothing. For all three resulting models, we chose parameters jointly over the model parameters and \pes\ parameters
to maximize validation set accuracy, measured in $R^2$, the percent of label variation explained by the predictions.

Figure~\ref{fig:table_1_barchart} shows the holdout test set performance of smoothed and unsmoothed models for the three machine learning algorithms. Training, validation, and test sets are of size $n=20,000$ each. We performed this experiment over 10 random data draws. Smoothing is performed with respect to just the predictions, as well as with respect to the concatenated training set labels and test set predictions (validated on training set labels and validation set predictions).
See Appendix~\ref{sec:appendix_experiment_details} for details on hyperparameter settings.  

For all three methods, applying \pes\ with the test set predictions improves accuracy over the original predictions. Smoothing with the training points, in the spirit of semi-supervised learning, boosts accuracy further, as we might expect since there is no estimation error for the training labels.
Figure~\ref{fig:returns_to_training_n} in Appendix~\ref{sec:vary_training_n} shows a similar trend holds across sample set sizes ($n$).

\begin{figure*}[!h]
\centering{}
      \includegraphics[width=.9\textwidth]{\figpath 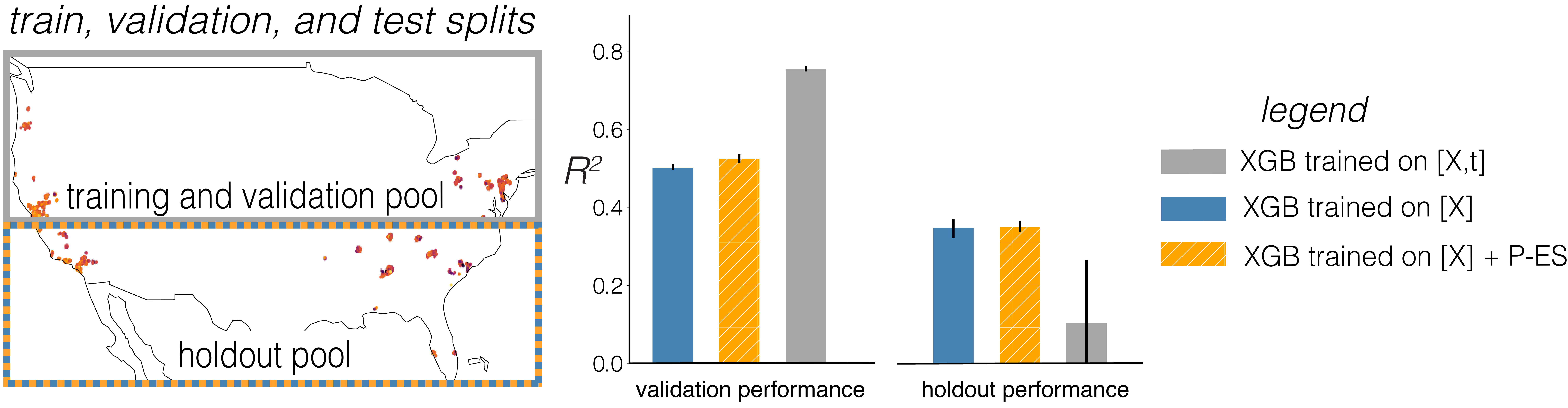}      \caption{In-sample (unsmoothed) validation performance and out-of-sample holdout performance for different methods of incorporating spatial index variables. Lines show min and max over 10 trials.}
      \label{fig:geo_extrap}
\end{figure*}

Table~\ref{table:zillow_compared_to_ssl} shows a comparison to alternative methods for reducing variance or inducing spatial consistency: kernel smoothing based only on the training set labels (without predictions), Gaussian process regression (GPR), Laplacian regularized least squares (LapRLS), a variance reducing shrinkage estimator ($S = \delta \cdot (\tfrac{1}{n} \vec{1}\vec{1}^\top) + (1-\delta)I$), and Gaussian harmonic energy minimization (HEM) (see Sec.~\ref{sec:related_work}). 

% The time reported for LapRLS is to solve the regularized least squares problem (LapRLS does not take $\hat{y}_{\textrm{XGB}}$ as input). The time for PES and HEM excludes learning $\hat{y}_{\textrm{XGB}}$ (an additive 8.1 seconds to both). 

Timing results underscore that \pes\ is a fast way to incorporate spatial structure (it incurs an $\bigO(n^2)$ 
%or $\bigO(n_{\textrm{val}}(n_{\textrm{val}}+n_{\textrm{train}}))$ 
additional runtime as opposed to $\bigO(n^3)$ for GPR, LapRLS, and HEM). 
The high variance in performance of kernel smoothing alone
%is likely due to 
may be explained by inherent difficulties in choosing hyperparameters in semi-supervised settings, as discussed by \cite{oliver2018realistic}. 
Accuracy of post-processing with \pes\ 
%in \Cref{table:zillow_compared_to_ssl}
is within 1.2 standard deviations
 %is on par with, but slightly worse than that 
of the HEM method %which performs a more complicated smoothing operator, 
which takes roughly $30\times$ as long to run in this instance over the chosen set of hyperparameters (see Appendix~\ref{sec:appendix_experiment_details}). 

Runtimes for post-processing procedures are reported as the additional time compared to not running the post-processing procedure (on average, computing \pes\ predictions takes $27$ seconds on top of the $7.8$ seconds to run XGB over multiple hyperparameter configurations). 
The runtime numbers reported in~\Cref{table:zillow_compared_to_ssl} are for solving the exact HEM and LapRLS problems, using the inverse and with as much shared computation as possible.
We omit experimental comparison to LGC \citep{zhou2004learning,zhou2004ranking} as the adaption from multi-class classification and ranking problems to regression problems is nontrivial, but we note that it is an iterative metho where each iteration is $\bigO(n^2)$. The first iteration of the LGC algorithm is very similar to P-ES, so that similarity of accuracy of HEM and \pes\ with the smoothing matrix defined in Eq.~\eqref{eq:S_NW} suggests that a one-iteration approximation of these algorithms can be sufficient in some cases.

The last line in table~\Cref{table:zillow_compared_to_ssl} confirms that we get a very small increase in accuracy by smoothing the best SSL method; this is consistent with our understanding that this smoothing operator acts similarly to the graph-defined HEM operator. The reduction in computation to apply \pes\ significantly reduces the barrier to comparing to this family of algorithms as a baseline. Additionally, the computational speed of \pes\ makes it much easier to explore different choices of the weight matrix $W(t;\sigma)$ from which to form the smoothing matrix;  we consider this future work since such a matrix will likely be domain-specific.

We conclude our experiments with a final example of the danger of incorporating index variables with unique characteristics as predictive features. We compare the out-of-sample generalizability  of a model $\pred{y}(x,t)$ trained on the concatenated set of home attributes and geographic location as features with that of the model from previous experiments $\pred{y}(x)$ trained only on attributes with smoothing applied. 

Figure~\ref{fig:geo_extrap} shows the results of these two approaches for a spatial extrapolation experiment, where a training set and validation set of size $20,000$ each are sampled from the northern U.S., and a holdout set of size $20,000$ is sampled from a disjoint southern segment.\footnote{A figure with all possible data pipelines including smoothing $[X,t]$ is given in Appendix~\ref{sec:appendix_extrapolation_figure}.}

We might expect that incorporating latitude and longitude as features could cause the first model to overfit, whereas distribution shift in neighborhood structure might negatively impact the learning of parameters for \pes.
Indeed, while the XGB predictor trained with locations and attributes (XGB trained on $[X,t]$) has better validation performance than the location-agnostic predictor (XGB trained on $[X]$), it performs much worse on the holdout set ($0.10$ vs $0.35$ average $R^2$). However, applying \pes\ does not degrade performance ($0.002$ average $R^2$ increase). 
In this scenario, the distribution shift with respect to $t$ is so severe that it is possible to overfit by using $t$ as a feature. Incorporating structure in $t$ with $\pes$, on the other hand, is much more robust to this distribution shift.

  % \subsection{Predicting housing price from imagery.}
  %   \inputtex{zillow_exp_imgs}

\vspace{-.5em}

\section{CONCLUSION }
\vspace{-.5em}
\label{sec:conclustion}
%!TEX root = ../aistats_draft/pes_aistats.tex

We introduce post-estimation smoothing as a method for incorporating %often-overlooked 
structural indices like time or location as valuable information sources in machine learning predictions. %Using these index variables to perform local variance reduction of a priori learned predictors has the benefits of (1) explicitly encoding that index variables are different from standard `features,' and (2) computational efficiency of never needing to retrain the original predictor.  
%We showed that a simple kernel suffices to elicit and exploit this structure for large scale temporal and geo-spatial tasks. 
Theory and experiments underscore that \pes\ is an effective and robust way to incorporate structured index variables in prediction, at much less cost than traditional semi-supervised methods.

The performance of \pes\ depends on the accuracy of the original predictions. If predictions $\hat{y}$ are very far from y, smoothing is unlikely to remedy this. While decoupling smoothing from the original prediction may be limiting, we have shown that it can be advantageous when viewing \pes\ as a diagnostic method or a baseline with which to compare more complex methods.

This work opens a door for extensions to applications where index variables satisfy less physical notions of distance (e.g., word embeddings), and to analysis characterizing when decoupling local consistency and prediction can be close to the optimal integrated approach. 
Future work could also consider multivariate labels $y$ with correlation among their elements.
%, and integrate this correlation into a \pes\ operation. 
%
Lastly, it will likely be worthwhile to 
%go beyond the Nadaraya-Watson estimator and
investigate more structured weight matrices $W$ (other than the Nadaraya-Watson estimator) which form the basis of the smoothing matrix $S$. 
Due to the decoupling of smoothing from the original prediction, practitioners can try  domain-specific weight matrices with marginal extra cost.  

When the goal is to obtain accurate predictors, no data should be overlooked. However, index variables such as time and space should be incorporated with care. We propose that post-processing is a natural and effective way to utilize this structure, and show it is robust to different tasks, predictors, and sampling patterns.
% We hope that these findings spark more work on characterizing different data sources and their innate structures, both in theory and practice. 

\small{
\subsubsection*{Acknowledgements}
\vspace{-.5em}
We thank Solomon Hsiang for thoughtful discussions on this topic. 
We also thank the authors of~\cite{kanazawa2019humanDynamics}, and in particular Jason Zhang, for their support in replicating their pose estimation predictions.
This material is based upon work supported by the NSF Graduate Research Fellowship under Grant No. DGE 1752814.}

\pagebreak
\bibliographystyle{plainnat}
\bibliography{pes_arxiv.bib}

\clearpage

\onecolumn

\appendix
\section{Appendix to Post-Estimation Smoothing: A Simple Baseline for Learning with Side Information}
  \subsection{On satisfying the conditions of \Cref{claim:smoothing_helps} ($\gamma + \beta <1$)
    \label{sec:gamma_beta_conditons}}
    Recall the definitions $\gamma(\err, W)$ and $\beta(\err, W; y)$:
\begin{align*}
\gamma(\err,W) &:= \EE[\err^\top W \err]/\EE[ \|\err\|_2^2] \\
\beta(\err,W; y) &:= {\EE[\err^\top (W-I)y ]}/{\EE\left[\|\err\|_2^2\right]}~.
\end{align*}
The condition $\gamma + \beta < 1$ captures a trade-off between choosing a weight matrix $W$ which reduces the magnitude of the errors (small $\gamma$), while not affecting too much the signal in the predictions (small $\beta$). The next paragraph shows that under a reasonable assumption on the predictions, $\beta + \gamma < 1$ can always be satisfied. The paragraph after details practical considerations in picking W and checking the conditions of the theorem.

Manipulation of the definitions of gamma and beta shows that the condition $\beta + \gamma < 1$ is equivalent to the condition $\EE[\err^\top(W-I) \hat{y}] <0$, which is always satisfiable with some $W$, so long as $\EE[\err\hat{y}^\top]$ is not the all zeros matrix. Further, when $|\EE[\err^\top y]| < \EE[\err^\top \err]$, $W = t\cdot I$ for any $t < 1$ will suffice so that $\beta + \gamma < 1$. The wide range of possible $t$ is because the matrix $W$ is combined in a convex combination with the identity matrix to form $S_c(t)$ in Eq.~\eqref{eq:S_c}.

\Cref{claim:unconstrained_opt} and \Cref{exmp:linear} show that an optimal smoothing matrix averages out errors in the predictions, depending on the structure in $y$ and $\err$. We’d like our empirical choice of $W$ to be close to this optimal matrix. For practical applications, we could (a) use empirical covariance matrices from training/validation data to inform our choice of W, and/or (b) for a pre-specified $W$ we could estimate $\gamma$ and $\beta$ by using the training/validation data to estimate $\err$ and $y$. We suspect that estimating $\gamma$ and $\beta$ in this way may not be practically necessary, for the following reason. If the chosen matrix $W$ does not reduce the mean squared error for any choice of $c \in (0,1]$, then cross validation over parameter $c$ will result in $c=0$, such that no smoothing occurs. Since cross-validating over $c$ amounts to only vector (not matrix) operations, it is practical to sweep over a large number of possible $c$’s. Thus, it could be just as fast to check if smoothing with matrix $S_c$ (for any of the $c$’s) reduces the MSE as to check the condition $\gamma+\beta < 1$.

  \subsection{Proof of~\Cref{claim:smoothing_helps}}
    \label{sec:proof_of_thm1}
    %!TEX root = ../neurips_draft/pes_neurips.tex

We now prove~\Cref{claim:smoothing_helps} in full generality. Recall the original theorem statement:
\smoothinghelps*
\begin{proof}
Let $\mu := \EE[\err] = \EE[\hat{y} - y]$.
The squared error ($n~\times $ the MSE) of using smoothing matrix $S_c = c W + (1-c)I$ decomposes as:
\begin{align*}
\|S_c \hat{y} - y\|_2^2 &= \|c (W\hat{y} - y) + (1-c)(\hat{y}-y)\|_2^2 \\
 &= \|c (W\hat{y} - y) + (1-c)(\err)\|_2^2 \\
&= 
c^2 \|W\hat{y} - y\|_2^2 + (1-c)^2 \| \err\|_2^2 
+ 2c(1-c) 
(\err^\top W\err + \err^\top(W-I)y) 
\end{align*}
so that the expected reduction in MSE is given by
 \begin{align*}
\EE \left[\|S_c \hat{y} - y\|_2^2\right] - \EE \left[\|\hat{y} - y\|_2^2\right]&= 
c^2 \EE \left[\|W\hat{y} - y\|_2^2 \right]
+ \left(1 + (c^2 - 2c) + 2(c-c^2)\gamma\right) \EE \left[\| \err\|_2^2  \right] \\
& \quad \quad + 2c(1-c) \EE \left[\err^\top(W-I)y)\right]  - \EE \left[\|\err\|_2^2 \right] \\
% &= 
% c^2 \EE \left[\|W\hat{y} - y\|_2^2 \right]
% + \left((c^2 - 2c) + 2(c-c^2)\gamma\right) \EE \left[\| \err\|_2^2 \right]  \\
% &\quad \quad + 2c(1-c)\mu^\top(W-I)y) \\
&= 
c^2 \EE \left[\|W\hat{y} - y\|_2^2 \right]
+ \left((c^2 - 2c) + 2(c-c^2)(\gamma+\beta)\right) \EE \left[\| \err\|_2^2 \right] 
 \end{align*}
This is a quadratic in $c$:
 \begin{align*}
\EE \left[\|S_c \hat{y} - y\|_2^2\right] - \EE \left[\|\hat{y} - y\|_2^2\right]
& = c^2 \left( \EE\left[\|W\hat{y} - y\|_2^2 \right] + (1-2(\gamma+\beta))\EE \left[\| \err\|_2^2 \right] \right) \\
&\quad\quad  + 2c \left((\gamma + \beta -1)\EE \left[\| \err\|_2^2 \right] \right) 
 \end{align*}
We first show that under the assumptions above, the above expression is convex. Afterwards, we will show that the nonzero root is strictly greater than zero, and therefore conclude that there must be a value $c \in (0,1]$ for which the objective is negative. We first get a handle on the coefficient of the quadratic term:
\begin{align*}
\EE\left[\|W\hat{y} - y\|_2^2 + (1-2(\gamma+\beta)) \| \err\|_2^2  \right] 
&= 
\EE\left[\|W\hat{y} - y\|_2^2+ \| \err\|_2^2 - 2 \err^\top W \err  + 2 \err^\top (I - W)y \right] \\
%& = \EE\left[\|W\hat{y} \|_2^2 - 2 y^\top W \pred{y} + \|y\|_2^2 + \| y\|_2^2  -2 \pred{y}^\top y + \| \pred{y}\|_2^2 - 2 \err^\top W (\err+y) + 2 \err^\top y \right] \\
%& = \EE\left[\|W\hat{y} \|_2^2 - 2 y^\top W \pred{y} + 2\|y\|_2^2   -2 \pred{y}^\top y + \| \pred{y}\|_2^2 - 2 \err^\top W \pred{y} +  2 \pred{y}^\top y - 2 \|y\|_2^2 \right] \\
& = \EE\left[\|W\hat{y} \|_2^2 - 2 y^\top W \pred{y}  + \| \pred{y}\|_2^2 - 2 \err^\top W \pred{y}   \right] \\
%& = \EE\left[\|W\hat{y} \|_2^2 - 2 \pred{y}^\top W \pred{y}  + \| \pred{y}\|_2^2  \right] \\
& = \EE\left[\|(W-I)\hat{y} \|_2^2 \right] \\
&\geq 0
\end{align*}
The coefficient on the quadratic term is nonnegative, so that the expression is convex in $c$. Now we show that under the conditions outlined in the theorem statement, the coefficient on the linear term is negative. Recall the condition that the matrix $W$ acts close to the identity on $y$ but close to the zero matrix on $\err$, with respect to the errors: $ \gamma(\err,W) + \beta(\err,W; y) < 1$. When this conditions holds, we have
\begin{align*}
2 \left((\gamma + \beta -1)\EE \left[\| \err\|_2^2 \right]\right) < 0 ~.
\end{align*} 
Thus, the optimal $c$ value is given as
\begin{align*}
c^* = 
\frac{\left(1 - (\gamma + \beta)\right)\EE \left[\| \err\|_2^2 \right]}{\EE\left[\|W\hat{y} - y\|_2^2 \right] + (1-2\gamma - 2 \beta) \EE\left[\| \err\|_2^2\right] }~.
\end{align*}
Since $c^*$ is always positive, by convexity and continuity of the objective function, the optimal value for $c$ within the range $(0,1]$ is $\min(c^*, 1)$. 

If $c^* >1$, this implies that 
\begin{align*}
\left(1 - (\gamma + \beta)\right)\EE \left[\| \err\|_2^2 \right] &> \EE\left[\|W\hat{y} - y\|_2^2 \right] + (1-2\gamma - 2 \beta) \EE\left[\| \err\|_2^2\right] \\
(\gamma + \beta))\EE \left[\| \err\|_2^2 \right] &> \EE\left[\|W\hat{y} - y\|_2^2 \right]~.
\end{align*}
If this is the case, then clipping the chosen $c$ to be $c=1$ (denote the resulting smoothing matrix $S_1$) will result in expected MSE decrease
\begin{align*}
\EE \left[\tfrac{1}{n}\|S_{1}\hat{y} - y\|_2^2\right] - \EE \left[\tfrac{1}{n}\|\err\|_2^2\right]
& = \EE \left[\tfrac{1}{n}\|W\hat{y} - y\|_2^2\right] - \EE \left[\tfrac{1}{n}\|\err\|_2^2\right]
\\
< - (1 - \gamma - \beta)\EE \left[\tfrac{1}{n}\|\err\|_2^2\right]~.
\end{align*}
Otherwise (if $c^* \leq 1$), the resulting expected MSE decrease is upper bounded as
\begin{align*}
\EE \left[\tfrac{1}{n}\|S_{c^*}\hat{y} - y\|_2^2\right] - \EE \left[\tfrac{1}{n}\|\err\|_2^2\right]
& \leq 
- \frac{\left(1 - \gamma - \beta\right)^2\EE \left[\| \err\|_2^2 \right]^2}{n \left(\EE\left[\|W\hat{y} - y\|_2^2 \right] + (1-2\gamma - 2 \beta) \EE\left[\| \err\|_2^2\right] \right) } \\
& = 
- (1-\gamma - \beta)\EE \left[\tfrac{1}{n}\| \err\|_2^2 \right] \cdot \frac{\left(1 - \gamma - \beta)\right)\EE \left[\| \err\|_2^2 \right]}{\left(\EE\left[\|W\hat{y} - y\|_2^2 \right] + (1-2\gamma - 2 \beta) \EE\left[\| \err\|_2^2\right] \right)}~.
\end{align*}
The optimal resulting MSE reduction from using $S_{c}$ where $c = \min\{c^*,1\}$ is then bounded as
\begin{align*}
&\EE \left[\tfrac{1}{n}\|S_{c}\hat{y} - y\|_2^2\right] - \EE \left[\tfrac{1}{n}\|\hat{y} - y\|_2^2\right] \\
&\quad\quad\quad \leq - (1 - \gamma - \beta) \EE\left[\tfrac{1}{n}\| \err\|_2^2\right] \cdot \min \left\{ 1,
\frac{(1-\gamma-\beta)\EE\left[\| \err\|_2^2\right]}{\left(\EE\left[\|W\hat{y} - y\|_2^2 \right] + (1-2\gamma-2\beta)\EE\left[\| \err\|_2^2\right]\right)}\right\} \\
&\quad\quad\quad < 0~. \qedhere
\end{align*}
\end{proof}

    \subsection{Proof of~\Cref{claim:unconstrained_opt}}
    \label{sec:proof_of_lem1}
    %!TEX root = ../neurips_draft/pes_neurips.tex

We now provide a proof of \Cref{claim:unconstrained_opt}. Recall the original statement:
\unconstrainedopt*
\begin{proof}
Setting the matrix differential of the following convex objective to zero, any solution $S^*$ to
\begin{align*}
S^* &= \argmin_{S \in \RR^{n \times n}} \EE\left[\tfrac{1}{n}\| S\pred{y} - y \|_2^2\right] 
\end{align*}
satisfies
\begin{align*}
\frac{\partial}{\partial S} \EE\left[ (S \pred{y} - y )^\top (S\pred{y} - y ) \right]
%&=2\EE\left[  S (\pred{y} \pred{y}^\top) - (y \pred{y}^\top ) \right]\\
& =  2 (S K_{\pred{y} \pred{y}} - K_{y \pred{y}}) = 0~.
\end{align*}
If $K_{\pred{y} \pred{y}}$ is positive definite (and thus invertible), the objective is strictly convex and the unique optimal solution is
\begin{align*}
S^* & = K_{y \pred{y}} (K_{\pred{y}\pred{y}})^{-1 } \\
 &= I-K_{\err \pred{y}} (K_{\pred{y}\pred{y}})^{-1 } 
\\
 &=I - (K_{\err\err} +K_{\err y})(K_{y y} +K_{y\err} + K_{\err y} + K_{\err\err})^{-1 } ~.
\end{align*}
since the identity matrix $I$ is within the set of possible estimators ($\RR^{n \times n}$), we know that the resulting objective satisfies $ \EE\left[\| S^* \pred{y} - y \|_2^2\right] \leq  \EE\left[\|\pred{y} - y \|_2^2\right]$. In fact, applying properties of the trace operator (cyclic property, invariance to transposes) gives the following expression for the reduction in expected squared error:
\begin{align*}
\EE\left[\|\pred{y} - y \|_2^2 - \| S^* \pred{y} - y \|_2^2\right]
% &=\tr{
%  K_{yy} + K_{\pred{y}\pred{y}} - 2K_{y\pred{y}} -
%  \left( K_{yy} + 
% - 2 S^* K_{\pred{y}y} + (S^*)^\top S^* K_{\pred{y}\pred{y}}
% \right)
%   }
%  \\
&=\tr{
K_{yy} + K_{\pred{y}\pred{y}} - 2K_{y\pred{y}} - K_{yy} + K_{y \pred{y}} \left(K_{\pred{y}\pred{y}}\right)^{-1}K_{\pred{y} y}}
\\
%  &=\tr{
%   K_{\pred{y}\pred{y}} - 2K_{y\pred{y}} + K_{y \pred{y}} \left(K_{\pred{y}\pred{y}}\right)^{-1}K_{\pred{y} y}}
%  \\
 &=\tr{
 (K_{\pred{y}\pred{y}} - K_{y \pred{y}} ) (K_{\pred{y}\pred{y}})^{-1}(K_{\pred{y}\pred{y}} - K_{\pred{y}y })}
 \\
  % &= \tr{(K_{\err \pred{y}})(K_{\err \pred{y}} + K_{y \pred{y}})^{-1}(K_{ \pred{y}\err})} \\
 &= \tr{(K_{\err\err} + K_{y\err })^\top (K_{yy} + K_{\err\err} + K_{\err y} + K_{y \err})^{-1}(K_{\err\err} + K_{y\err })} ~.
\end{align*}
Applying a matrix trace inequality for positive definite matrix $A$ and positive semi-definite matrix $B$: $\tr{A^{-1}B}\geq \lambdamin(A^{-1}) \tr{B} = \tr{B}/\lambdamax{(A)} \geq  \tr{B}/\tr{A}$ gives an upper bound on the reduction:
\begin{align*}
\EE\left[\|\pred{y} - y \|_2^2 - \| S^* \pred{y} - y \|_2^2\right]
 \geq \frac{\tr{\left(K_{\err\err} + K_{y\err }\right)\left(K_{\err\err} + K_{y\err }\right)^\top}}{\tr{K_{yy} + K_{\err\err} + K_{\err y} + K_{y \err}}}~.
\end{align*}

 Note that $(K_{\err\err} + K_{y \err})(K_{\err\err} + K_{y \err})^\top$  and $K_{yy} + K_{\err\err} + K_{\err y} + K_{y \err} = K_{\hat{y}\hat{y}}$ are positive semi-definite by construction and positive definite by assumption, respectively.  
\end{proof}

    \subsection{Linear example (continued)}
      \label{sec:appendix_example}
      % !TEX root=../pes_arxiv.tex

Here we give a more thorough analysis of the example presented in example~\ref{exmp:linear} in the main text. 
Recall the setting: the zero-mean stochastic processes $x(t)$ and $y(t)$ which are dependent on a third zero-mean hidden process $z(t)$, but with independent additive Gaussian noise $\omega(t)$, $\mu(t)$, respectively. In particular:
\begin{align*}
%\label{eq:linear_example_z}
z &\sim \mathcal{N}(0, \Sigma_z) \\
%\label{eq:linear_example_x}
x(t) &= z(t) + \omega(t), \quad \omega(t) \sim_{i.i.d.} \mathcal{N}(0,\sigma^2_x) \\
%\label{eq:linear_example_y}
y(t) &= c \cdot z(t) + \mu(t), \quad \mu(t) \sim_{i.i.d.} \mathcal{N}(0,\sigma^2_y) 
\end{align*}
The autocorrelation matrices show that there is shared variation due to the ``hidden" process $z$:
\begin{align*}
K_{xx}[t,s] &= K_{zz}[t,s] + K_{\omega \omega}[t,s] \\ 
K_{yy}[t,s] &= c^2 K_{zz}[t,s] + K_{\mu \mu }[t,s]
\\
K_{xy}[t,s] &= c K_{zz}[t,s]
\end{align*}
Consider the problem of learning a predictor for unseen samples by learning the 1-dimensional regression weight $\hat{c}$ from a sample of $\{x_i,y_i\}_{i=1}^n$ data pairs drawn from the distribution above.
Then for a fresh, independently drawn sample will have predicted value $\pred{y} = \hat{c}\cdot x$, and 
\begin{align*}
K_{y\pred{y}} = \EE\left[ y \pred{y}^\top\right] = \EE \left[ y (\hat{c}x)^\top \right] = \EE[\hat{c}]  K_{xy}^\top
= c \EE[\hat{c}] K_{zz}~.
\end{align*}
Similarly,
\begin{align*}
K_{\pred{y}\pred{y}} = \EE \left[ \hat{c}x (\hat{c}x)^\top \right]= \EE[\hat{c}^2](K_{zz}+K_{\omega \omega})
= \EE[\pred{c}^2](K_{zz}+\sigma_x^2 I)~.
\end{align*}
Then the optimal smoothing matrix from \Cref{claim:unconstrained_opt} is 
\begin{align*}
S^* = 
K_{y\pred{y}}\left(K_{\pred{y}\pred{y}}\right)^{-1} = \frac{c\EE[\hat{c}]}{\EE[\hat{c}^2]} K_{zz} 
\left( K_{zz} + \sigma_x^2I\right)^{-1}~.
\end{align*}

 The model defined above can be described as an ``errors in variables" model, if we consider $z$ as the true regresssor and $x$ as an error-imbued observation of it. Under such a model, total least squares provides a consistent estimator of $c$ (see below), and thus it is the estimator that we analyze in the main text. However, we are concerned first and foremost with recovering $y$ without postprocessing, the ordinary least squares estimator might be a preferable solution. We first expand upon the exposition from the main paper of the example under the total least squares estimator, then follow with a discussion of using the ordinary least squares estimator in this context. 

\paragraph{Total least squares (TLS) estimator.}
To compute the forms of the auto-correlation matrices above for the TLS estimator, we make use of the following fact found, for example,  in \cite{sabinevanhuffel1991,schneeweiss1976consistent}:
\begin{itemize}
% \item {For errors in variables models such as the one we consider here, the TLS estimator $\hat{c}_{tls}$ is unbiased, so that as $n \rightarrow \infty$
\item{For the errors in variables model described above, the asymptotic distribution of the TLS estimator is normal, with mean $c$, and variance approaching 0 as $n \rightarrow \infty$. %Further, the finite sample variance can be estimated as:
% \begin{align*}
% %todo{\textrm{put in finite sample results from that paper}}
% \end{align*}
}
\end{itemize}
Which gives us the approximations  approximations
$\EE\left[ \hat{c}_{tls}\right]\approx c$, 
and $\EE[\hat{c}]^2 \approx \EE[\hat{c}^2]$.
The calculations in the main text are thus written out more expositionally as:\\ 
Expected unsmoothed performance:
\begin{align*}
\EE\left[\tfrac{1}{n}\|\pred{y} - y\|_2^2\right]
& = \tfrac{1}{n}\tr{ K_{yy} - 2K_{y\pred{y}} + K_{\pred{y}\pred{y}}} \\
& = \tfrac{1}{n}\tr{ 
c^2 K_{zz} + \sigma_y^2 I - 2 c \EE[\hat{c}] K_{zz}
+
\EE[\pred{c}^2](K_{zz}+\sigma_x^2 I)}  \\
% \end{align*} 
% Using the approximation $\EE[\hat{c}]^2 \approx \EE[\hat{c}^2]$, which holds as $n \rightarrow \infty$,
% \begin{align*}
% \EE\left[\frac{1}{n}\|\pred{y} - y\|_2^2\right]
&\approx  \sigma^2_y + c^2 \sigma^2_x ~.
\end{align*} 

From~\Cref{claim:unconstrained_opt}, the expected smoothed performance using $S^*$ is:
\begin{align*}
\EE\left[ \frac{1}{n}\|S^* \pred{y} - y\|_2^2\right]
 %& = \tfrac{1}{n}
 % \tr{
 % K_{yy} - K_{y \pred{y}} \left(K_{\pred{y}\pred{y}}\right)^{-1}K_{\pred{y} y}} \\
  & = \frac{1}{n} \tr{c^2 K_{zz} + \sigma^2_y I
 -
c^2 \frac{\EE[\hat{c}]^2}{\EE[\hat{c}^2]} K_{zz}^2 \left(K_{zz} + \sigma_x^2 I \right)^{-1} }  \\
&\approx \frac{c^2}{n}\tr{ K_{zz}\left(I - K_{zz}\left(K_{zz} + \sigma_x^2 I \right)^{-1}\right) }
+ \sigma^2_y  \\
& = c^2\sigma_x^2 \left(1 - 
\frac{1}{n} \tr{\left(\sigma_x^{-2}K_{zz} +I \right)^{-1} } \right) \
+ \sigma^2_y  \\
& \geq \sigma^2_y~.
\end{align*}

Using the second to last line above, the expected decrease in MSE achieved from applying the optimal linear smoothing matrix to the asymptotic total least squares estimator is then
\begin{align*}
\EE\left[ \frac{1}{n}\|\pred{y} - y\|_2^2\right]
-
\EE\left[ \frac{1}{n}\|S^* \pred{y} - y\|_2^2\right]
&\approx 
\frac{c^2\sigma_x^2}{n} \tr{\left(\sigma_x^{-2}K_{zz} +I \right)^{-1} } \\
&\geq 
\frac{c^2 \sigma_x^2}{n}\sum_n \frac{1}{1 + \sigma_x^2\cdot\lambdamax(K_{zz})} = c^2 \frac{\sigma_x^2}{1 + \sigma_x^2\cdot\lambdamax(K_{zz})}
\end{align*} 
where $\lambdamax(\cdot)$ denote the maximum eigenvalue of a matrix.

\iftoggle{arxiv}{
\begin{figure}[!t] %  figure placement: here, top, bottom, or page
   \centering
   \includegraphics[width=.9\columnwidth]{\figpath 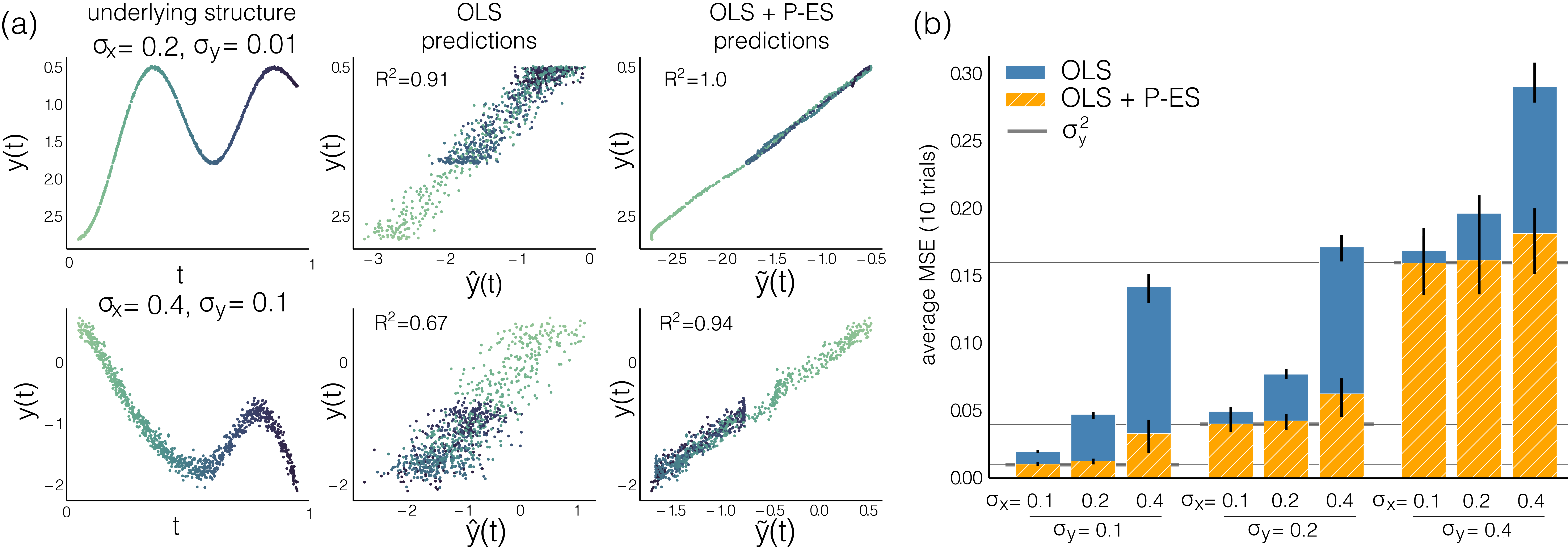}
      \caption{Figure for exact same run of simulations but using ordinary least squares (OLS) estimator instead of total least squares, as in Fig.~\ref{fig:simulation_panel_tls}.}
   \label{fig:simulation_panel_ols}
\end{figure}
}
{
\begin{figure}[!t] %  figure placement: here, top, bottom, or page
   \centering
   \includegraphics[width=.9\columnwidth]{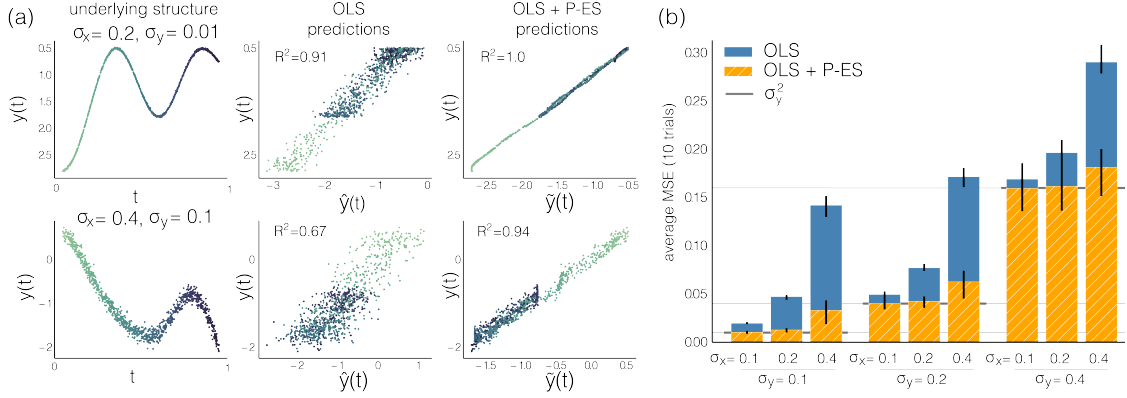}
      \caption{Figure for exact same run of simulations but using ordinary least squares (OLS) estimator instead of total least squares, as in Fig.~\ref{fig:simulation_panel_tls}.}
   \label{fig:simulation_panel_ols}
\end{figure}	
}

\paragraph{Ordinary Least Squares (OLS) estimator}
Due to the noise process in $x$, OLS will produce a biased estimator $\hat{c}$:
\begin{align*}
\hat{c}_{ols} &= (x^\top x)^{-1} x^\top y \\
&= ((z+\omega)^\top (z+\omega))^{-1} (z+\omega)^\top (cz + \mu) 
\end{align*}
$\mu$ is uncorrelated with $z$ and $\omega$, so that
\begin{align*}
\EE \left[\hat{c}_{ols}\right] 
%&= c\EE\left[\frac{(z^\top z + z^\top\omega)}{(z+\omega)^\top (z+\omega)}  \right] \\
&= c\left(1 - \EE\left[\frac{w^\top w +z^\top\omega}{(z+\omega)^\top (z+\omega)} \right] \right)~.
\end{align*}
As $n\rightarrow \infty$,
\iftoggle{arxiv}{}{\vspace{-2em}}
\begin{align*}
\EE \left[\hat{c}_{ols}\right]  \rightarrow c \left( 1 - \frac{\sigma_x^2}{\sigma_x^2 + \frac{1}{n}\tr{K_{zz}}} \right)~.
\end{align*} We see that the noise associated with $x$ biases the estimated regression coefficient to be shallower; this is a well known phenomenon in the errors-in-variables model termed attenuation bias. 
This bias limits the amount to which \pes\ can denoise the estimations, as shown in Fig.~\ref{fig:simulation_panel_ols}(B). In comparison to Fig.~\ref{fig:simulation_panel_tls}(B), we see that the unsmoothed OLS estimator exhibits the same qualitative behavior over the parameter selections as the unsmoothed TLS estimator. Moreover, the same pattern of the smoothed estimates (with performance floor around $\sigma^2_y$) is maintained in Fig.~\ref{fig:simulation_panel_ols}(B), although this is trend is less fitting for larger $\sigma_x^2$ (corresponding to larger magnitude of bias in $\hat{c}_{ols}$). 
%Now we need to wrangle $\EE \left[\hat{c}^2_{ols}\right]$:
% \begin{align*}
% \EE\left[ \frac{\left((z+\omega)^\top (c z + \mu)\right)^2}{\left((z+\omega)^\top (z+\omega)\right)^2} \right]
% = 
% \EE\left[ \frac{=}
% {\|z+\omega\|^4} \right]
% \end{align*}
% \todo{get a grip on the second order term.}

\section{Experiment Details}
      \label{sec:appendix_experiment_details}
      All experiments were run on a machine with 48 cores, each of them an Intel(R) Xeon(R) CPU E5-2670 v3 @ 2.30GHz, and 256G RAM. All experimental code is written in python, and the relevant libraries used are listed below. Our code is available at \url{\githuburl}. 
Instructions for downloading and using the intermediate video predictions from~\cite{kanazawa2019humanDynamics} are detailed there. The housing data is  provided by Zillow through the Zillow Transaction and Assessment Dataset (ZTRAX). More information on accessing the data can be found at \url{ http://www.zillow.com/ztrax}. (The results and opinions are those of the authors of this work and do not reflect the position of Zillow Group).

\subsection{Video experiments}
\paragraph{Metrics.}
For consistency, we use the same metrics as reported in \cite{kanazawa2019humanDynamics}, and the same code to calculate these metrics. All metrics are defined per video, and averaged over all videos. A description of each metric is given here; See \cite{kanazawa2019humanDynamics} and \url{https://github.com/cbsudux/Human-Pose-Estimation-101} \citep{Babu2019} for further explanation:
\begin{itemize}
\item{Percentage key points (PCK)}: percentage of 2D key points that fall within $\alpha\cdot \max\{ h, w\}$ of the labeled key point, where $h$ and $w$ parameters of a per-frame tight bounding box around the entire person; here $\alpha = 0.05$.
\item{Mean per joint position error (MPJPE)}: Mean euclidean distance of predicted to ground truth joint, averaged over joints in the human pose model (calculated after aligning root joints), measured in millimeters.
\item{Mean per joint position error after Procrustes alignment (PA-MPJPE)}: MPJPE after alignment to the ground truth by Procrustes alignment method, measured in millimeters.
\item{Acceleration Error (Accel Err)}: defined in \cite{kanazawa2019humanDynamics} as ``the average difference between ground truth 3D acceleration and predicted 3D acceleration of each joint in $mm/s^2$.''
\item{Acceleration (Accel)} For 2D datasets, measures ``acceleration in $mm/s^2$'' \citep{kanazawa2019humanDynamics}. Note that this metric is only useful in conjunction with other metrics, as a baseline constant predictor would achieve 0 acceleration. However, for predictions that also do well on PCK, lower acceleration is more meaningful.
\end{itemize}
\paragraph{Parameter tuning.} We started with a grid search of $\sigma \in [0.5,1,2,4]$ and $c \in [0.0,0.2,0.4,0.6,0.8,1.0]$ and then interpolated best values once to obtain this final set.  Specifically, this meant including $\sigma = 3$ and $c \in [0.5,0.7,0.9]$.

\subsection{Predicting house price from attributes}

\paragraph{Metrics.}
$R^2$, or coefficient of determination is a metric which reports the percent of squared deviation in the independent labels which is explained by the predictions. Formally it is defined as
\begin{align*}
R^2 = 1 - \frac{\sum_{i=1}^n (y_i - \pred{y}_i)^2}{\sum_{i=1}^n (y_i - \textrm{avg}(y))^2}.
\end{align*}  
It is possible that this score can be negative; in this case we clip negative $R^2$ values at zero in computing averages and ranges (in our experiments, this only occurs for spatial extrapolation when locations are considered as features). We used the implementation of $R^2$ available via $\mathtt{sklearn.metrics.r2\_score}$.

\paragraph{Dataset.}
The Zillow Transaction and Assessment Database (ZTRAX) \citep{ztrax} contains home sales of many different types; we restricted our dataset to single family homes. Only the most recent sale for a property id and location was considered, and after that only home sales occur after the year 2010 (dated by the column `contract year'). Any observation for which any of the 12 features considered (listed below) were missing was dropped. The resulting dataset contains $608,959$ homes sales spread across the United States. 

Features included were: year built (from 2010), number of stories, number of rooms, number of bedrooms, number of baths, number of partial baths, size (sqft), whether there is heating, whether there is air conditioning, the contract year, the contract month, and whether the home was new; location was encoded as the latitude and longitude of the home, and target label is the most recent sale price of the home.

Hyperparameters searched for generating Fig.~\ref{fig:table_1_barchart} are given in~\Cref{table:fig2_hps}. The ten random trials for the experiments in Fig.~\ref{fig:table_1_barchart} where done as follows. For each trial we drew 60,000 data points from the total dataset, with replacement between trials. Then for each random draw, we allocated 20,000 points to the training, validation and test sets, such that no points were overlapping within in each trial. In each trial, hyperparameters were chosen to maximize validation performance for that single trial, and then the optimal hyperparameters defined the model that we applied to the holdout set. 

For the timing and methodological comparisons in~\Cref{table:zillow_compared_to_ssl}, we followed a similar procedure, but with only 10,000 points for training, validation and test sets, so that the total run times were reasonable and we could run enough trials to get a notion of variability in results. The parameters considered in this experiment and  the total number of parameter configurations swept over for each algorithm, are given in~\Cref{table:table2_hps}. 

The spatial extrapolation experiments followed the same sampling protocol as above for each trial, with the exception that training and validation sets were drawn from a pool of observations which lay above $37\degree$ in latitude ($376,615$ total observations), and the holdout sets were drawn from the remaining $232,344$ observations. For this experiment we considered hyperparameters $\mathtt{max\_depth} \in [5,10]$, $\mathtt{num\_estimators} \in [100,200]$, $\sigma \in \mathtt{logspace(-4,2,base=10,num=9)}$ and $c \in \mathtt{linspace(0,1.0,num=11)}$.

We used the existing $\mathtt{sklearn}$ implementation of $\mathtt{GaussianProcessRegressor}$ for GPR\footnote{Documentation available at:  \url{https://scikit-learn.org/stable/modules/generated/sklearn.gaussian_process.GaussianProcessRegressor.html}},
$\mathtt{xgboost.XGBRegressor}$ for xgboost\footnote{Documentation available at:
\url{https://xgboost.readthedocs.io/en/latest/python/python_api.html}},
and our own implementation for HEM and LapRLS which pre-computes the Gram matrix for efficiency (all code available at \githuburl). We also used our own implementation of the random features algorithm of~\cite{rahimi2009weighted}, so that each random feature is generated as a transformation of the original features $x$: 
\begin{align*}
\cos ( w^\top x + b);
 \quad w \sim \mathcal{N}(0, \sigma_{\mathtt{RF}}^2) 
 \quad b \sim \textrm{unif}(0, 2\pi)~.
\end{align*}
\begin{table}[ht]
  \caption{Hyperparameters considered in runs for~\Cref{fig:table_1_barchart}.}
   \label{table:fig2_hps}
  \centering
  \begin{tabular}{lll}
    \toprule
     method& hyperparameters considered \\
   \hline
Ridge Regression & $\lambda_{\mathtt{RR}} \in \mathtt{logspace(-6,4,base=10,num=5)}$\\
\hline
Random Features & number of random features $\in [100,200]$\\
 & $\sigma_{\mathtt{RF}} \in \mathtt{logspace(-8,-4,base=10,num=3)}$\\
 & $\lambda_{\mathtt{RR}} \in \mathtt{logspace(-6,-4,base=10,num=3)}$\\
\hline
XGB & $\mathtt{max\_depths} \in [2,5,10]$\\
& $\mathtt{num\_estimators} \in [100,200]$ & \\
 \hline
PES &  $\sigma \in \mathtt{logspace(-4,0,base=10,num=5)}$ \\
&  $c \in \mathtt{linspace(0,1.0,num=11)}$  \\
    \bottomrule
  \end{tabular}
\end{table}

\begin{table}[!ht]
  \caption{Hyperparameters considered in runs for~\Cref{table:zillow_compared_to_ssl}.}
   \label{table:table2_hps}
  \centering
  \begin{tabular}{lll}
    \toprule
     method& hyperparameters considered & total number of hyperparameters \\
    \hline
smoothing & $\sigma \in \mathtt{logpspace(-2,0,base=10,num=9)}$ & $9$ \\
\hline
XGB & $\mathtt{max\_depths} \in [2,5,10]$, & $ 6~(3 \times 2)$\\
& $\mathtt{num\_estimators} \in [100,200]$ & \\
\hline
+ shrinkage &  $\delta \in \mathtt{linspace(0,1,num=11)} $ & $66~(11 \times 6)$\\
 \hline
+ \pes &  $\sigma \in \mathtt{logspace(-4,0,base=10,num=5)}$ & $ 330~((5 \times 11) \times 6)$ \\
 & $c \in \mathtt{linspace(0,1,num=11)}$ & \\
\hline
LapRLS & $\lambda_\mathtt{ridge} \in \mathtt{logpspace(-2,4,base=10,num=5)}$,
& $25~(5 \times 5)$  \\
& $\lambda_\mathtt{lap} \in \mathtt{logpspace(-4,2,base=10,num=5)}$ & \\
\hline
GPR & $\alpha \in \mathtt{logspace(-6,0,num=3,base=10)}$ &  $48~(3\times4\times4)$\\
& $\sigma_\mathtt{const} \in \mathtt{logspace(-2,2,num=4,base=10)}$ & \\
& $\sigma_\mathtt{gpr} \in \mathtt{logspace(-2,2,num=4,base=10)}$ &  \\
\hline
HEM & $\sigma \in \mathtt{logspace(-4,0,base=10,num=5)}$ & $ 180~((5\times6)\times 6)$ \\
 & $\eta \in \mathtt{linspace(0.01,1,num=6)}$ & \\
    \bottomrule
  \end{tabular}
\end{table}

\subsection{Performance for different data set sizes \label{sec:vary_training_n}}
% !TEX root=../pes_arxiv.tex

\begin{figure*}[!h]
\centering
      \includegraphics[width=.9\textwidth]{\figpath 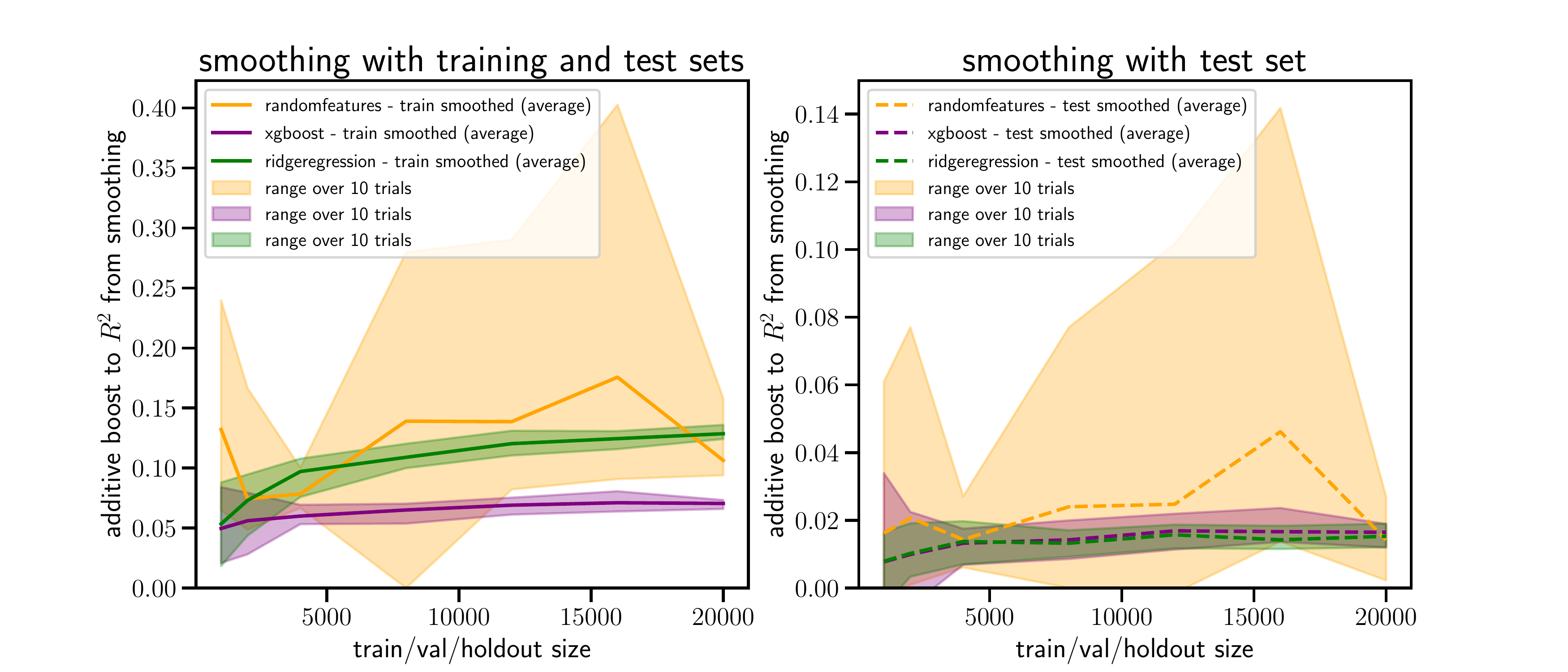}      \caption{Additive MSE increase due to smoothing for different training set sizes.}
      \label{fig:returns_to_training_n}
\end{figure*}

Here we study the accuracy increases from \pes\ as a function of the amount of training data. We varying train/validation and holdout set sizes in  $[1000,2000,4000,8000,12000,16000, 20000]$ and otherwise following the experimental setup in Figure~\ref{fig:table_1_barchart}). The resulting accuracy increases (from the unsmoothed predictions) for 10 random trials are shown in Figure~\ref{fig:returns_to_training_n}.

While the performance of \pes\ for the random features regressor is variable (due largely to the variability of the original predictor), we see for the other two predictors a trend that as the data sizes increase, the advantage to smoothing is not decreasing. With more training data, the underlying predictors capture better signal and there are more nearby predictions which with to smooth any given point, both of which are advantages for \pes.

\iftoggle{arxiv}{}{\vspace{12em}}
\subsection{Extrapolation experiments}
\label{sec:appendix_extrapolation_figure}
 \begin{figure}[h]%  figure placement: here, top, bottom, or page
 \centering
      \includegraphics[width=.8\textwidth]{\figpath 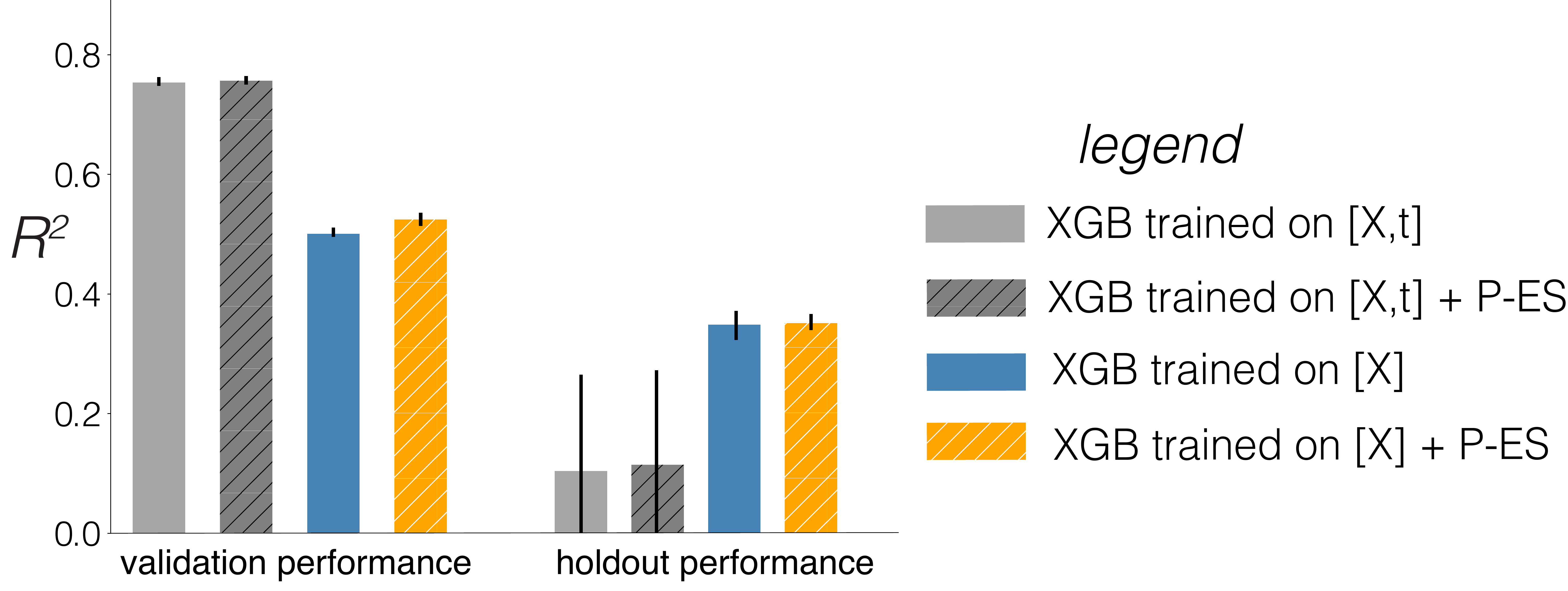}
      \caption{All comparisons for geographic extrapolation experiment.}
      \label{fig:geo_extrap_all_comparisons}   
\end{figure}

In Figure~\ref{fig:geo_extrap} of the main text, we compared two different methods for incorporating latitude and longitude in house price predictions. Figure~\ref{fig:geo_extrap_all_comparisons} shows the same plot, with the addition of smoothing on the predictions that included $t$ as features.
The aim of this experiment is to show that P-ES is robust to distribution shifts from e.g. extrapolation (not that it increases performance necessarily). Validation performance using [X,t] (left solid grey) is much higher than just using [X] (left blue), which might mislead a practitioner to include $t$ as a feature when in fact the holdout performance is much worse (right solid grey blue vs. right blue). In contrast, validation P-ES performance on [X] (left solid blue) is worse than including $t$ as a feature (left dashed grey), but exhibits no holdout degradation (left blue vs. right dashed grey).

\end{document}